\documentclass[pdflatex,sn-mathphys-num,a4paper]{sn-jnl}% Math and Physical Sciences Numbered Reference Style
\usepackage{graphicx}%
\usepackage{multirow}%
\usepackage{amsmath,amssymb,amsfonts}%
\usepackage{amsthm}%
\usepackage{mathrsfs}%
\usepackage[title]{appendix}%
\usepackage{xcolor}%
\usepackage{textcomp}%
\usepackage{manyfoot}%
\usepackage{booktabs}%
\usepackage[ruled,vlined]{algorithm2e}
\usepackage{listings}%
\usepackage{tikz}
\usepackage{tikz}
\usepackage{subcaption}
\usepackage{float}
\usepackage{placeins}
\usepackage{hyperref}
\usepackage{cleveref}
\usetikzlibrary{shapes,arrows,positioning}
\theoremstyle{thmstyleone}%
\theoremstyle{thmstyletwo}%

\theoremstyle{thmstylethree}%

\begin{document}

\title[Article Title]{\textbf{Organ-Agents: Virtual Human Physiology Simulator via LLMs}}


\author[1]{\fnm{Rihao} \sur{Chang}}
\equalcont{These authors contributed equally to this work.}
\author[1]{\fnm{He} \sur{Jiao}}
\equalcont{These authors contributed equally to this work.}
\author*[1]{\fnm{Weizhi} \sur{Nie}}\email{weizhinie@tju.edu.cn}
\author[1]{\fnm{Honglin} \sur{Guo}}
\author[3]{\fnm{Keliang} \sur{Xie}}
\author[4]{\fnm{Zhenhua} \sur{Wu}}
\author[3]{\fnm{Lina} \sur{Zhao}}
\author[4]{\fnm{Yunpeng} \sur{Bai}}
\author[1]{\fnm{Yongtao} \sur{Ma}}
\author[1]{\fnm{Lanjun} \sur{Wang}}
\author[1]{\fnm{Yuting} \sur{Su}}
\author[5]{\fnm{Xi} \sur{Gao}}
\author*[2,6]{\fnm{Weijie} \sur{Wang}}\email{weijie.wang@unitn.it}
\author[2]{\fnm{Nicu} \sur{Sebe}}
\author[6]{\fnm{Bruno} \sur{Lepri}}
\author*[5]{\fnm{Bingwei} \sur{Sun}}\email{sunbinwei@njmu.edu.cn}

\affil[1]{\orgname{Tianjin University}, \orgaddress{\city{Tianjin}, \country{China}}}

\affil[2]{\orgname{University of Trento}, \orgaddress{\city{Trento}, \country{Italy}}}
\affil[3]{\orgname{Tianjin Medical University General Hospital}, \orgaddress{\city{Tianjin}, \country{China}}}

\affil[4]{\orgname{Chest Hospital, Tianjin University}, \orgaddress{\city{Tianjin}, \country{China}}}
\affil[5]{\orgname{The Affiliated Suzhou Hospital of Nanjing Medical University}, \orgaddress{\city{Suzhou}, \state{Jiangsu}, \country{China}}}
%\affil[10]{\orgname{Fondazione Bruno Kessler}, \orgaddress{\city{Trento}, \country{Italy}}}
\affil[6]{\orgname{Fondazione Bruno Kessler}, \orgaddress{\city{Trento}, \country{Italy}}}

\abstract{
    Recent advances in large language models (LLMs) have enabled new possibilities in simulating complex physiological systems through reasoning, generation, and agentic coordination. In this work, we present Organ-Agents, a novel multi-agent framework that simulates the dynamics of human physiology using LLM-driven agents. Each agent, referred to as a Simulator, is assigned to model a specific physiological system such as the cardiovascular, renal, immune, or respiratory system. The training of the Simulators consists of two stages: supervised fine-tuning on system-specific time-series data, followed by reinforcement-guided inter-agent coordination that incorporates dynamic reference selection and error correction with assistantive agents. 
    To support training, we curated a cohort of 7,134 sepsis patients and 7,895 matched controls, constructing high-resolution, multi-domain trajectories covering 9 physiological systems and 125 clinical variables. Organ-Agents achieved high simulation accuracy on 4,509 held-out patients, with average per-system mean squared error (MSE) below 0.16 across all systems and robust performance across severity strata based on sequential organ failure assessment (SOFA) scores. Generalization capability was confirmed via external validation on 22,689 intensive care unit (ICU) patients from two tertiary hospitals, showing moderate performance degradation under distribution shifts while maintaining overall simulation stability. In terms of clinical plausibility, Organ-Agents reliably reproduces multi-system critical event chains (e.g., hypotension, hyperlactatemia, hypoxemia) with preserved event order, coherent phase progression, and minimal deviations in both trigger timing and physiological values. Subjective evaluation by 15 critical care physicians further confirmed the realism and physiological coherence of simulated trajectories, with mean Likert ratings of 3.9 and 3.7, respectively. The Simulator also supports counterfactual simulation under alternative fluid resuscitation strategies for sepsis, producing physiological trajectories and APACHE II scores that closely align with matched real-world patient groups. To further assess the preservation of clinically meaningful patterns, we evaluated Organ-Agents in downstream early warning tasks using seven representative classifiers. Most models showed only marginal AUROC degradation when transferring from real to generated and counterfactual trajectories, with performance drops generally within 0.04, indicating that the simulations preserved decision-relevant information for clinical risk simulation.
    Together, these results position Organ-Agents as a clinically credible, interpretable, and generalizable digital twin for in physiological modeling, enabling precision diagnosis, treatment simulation, and hypothesis testing across critical care settings. 
}




\maketitle

\section{INTRODUCTION}\label{sec1}
Human physiology operates as a continuously evolving, multi-organ system \cite{bashan2012network, lehnertz2020human}, yet clinical decision-making remains largely reliant on static, fragmented data. This mismatch causes critical healthcare shortcomings: 1) delayed detection of life-threatening events (e.g., sepsis); 2) poor model generalizability across physiological systems (e.g., cardiac models failing on kidney-induced heart failure); and 3) lack of interpretable clinical insights \cite{hama2025deep, xie2022deep}. Simulating human physiological functions offers a powerful solution to these challenges. By reconstructing the dynamic interplay between organ systems over time, such simulation models can enable proactive early warning, personalize treatment decisions based on patient-specific trends, and support plausible reasoning through counterfactual experimentation. 

However, even the most advanced existing approaches remain limited in their ability to simulate the real-time evolution of physiological systems. For instance, the Aitia Digital Twin model \cite{aitia2024causal} and Siemens Healthineers' Cardio Twin \cite{martinez2019cardio} have demonstrated the potential of organ-level mechanistic modeling. Yet, these approaches predominantly depend on predefined equations and static parameters, underpinned by extensive domain-specific priors including structural models and biophysical equations. While ensuring mechanistic fidelity, such rigid frameworks inherently limit their ability to adapt to dynamic physiological states and scale for whole-body real-time simulations. In parallel, several clinical models built on large language models (LLMs) have demonstrated strong performance in tasks such as retrospective classification, mortality simulation, and clinical document summarization. Representative examples include BioGPT \cite{luo2022biogpt}, trained on PubMed abstracts for biomedical text generation; ClinicalBERT \cite{huang2020clinicalbertmodelingclinicalnotes}, fine-tuned on electronic health records for downstream clinical tasks; and NYUTron \cite{jiang2023nyutron}, a transformer-based model developed for predictive analytics using both structured and unstructured hospital data. Nevertheless, these models predominantly focus on static representations of historical data and lack the temporal reasoning and inter-system coordination required for simulating continuous physiological processes. Most treat the diagnostic process as a static analysis of past observations, rather than as a dynamic inference grounded in evolving physiological mechanisms. They also lack the ability to simulate longitudinal trajectories under interventions, limiting their use in causal inference and scenario exploration. 

We posit that simulating human physiology requires a fundamentally different paradigm, one that supports time-resolved, adaptive, and faithful simulation. To advance this paradigm, three key challenges remain: 
(i) \textbf{Entangled interactions}: clinical monitoring produces dense, high-dimensional time-series data where numerous indicators shift asynchronously, obscuring inter-variable dependencies and temporal correlations; 
(ii) \textbf{State-aware dynamics}: organ interactions are nonlinear and evolve with changing physiological conditions, necessitating adaptable and context-aware inter-system coordination; 
(iii) \textbf{Physiological plausibilty}: simulations must not only fit observed trajectories but also adhere to underlying physiological mechanisms to ensure clinical validity. 

To holistically address the aforementioned challenges, we propose \textbf{Organ-Agents}, a unified, multi-agent simulation framework that models human physiology as a system of interacting, LLM-driven agents. By modularizing organ-specific dynamics while enabling adaptive inter-system coordination, Organ-Agents accommodates the asynchronous, nonlinear, and multivariate nature of clinical data and supports faithful, time-resolved simulation grounded in physiological mechanisms.

\begin{figure}[t]
\centering
\includegraphics[width=1\linewidth]{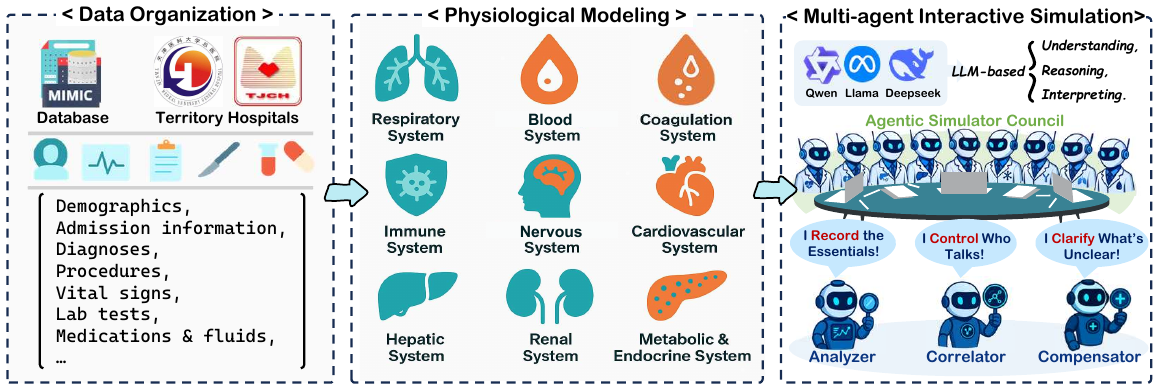}
\caption{Overview of Organ-Agents architecture. Organ-Agents integrates heterogeneous intensive care unit (ICU) data (including demographics, clinical notes, and 69 physiological indicators grouped by organ systems) from the MIMIC-IV dataset and multiple tertiary hospitals. Each physiological system is represented by an LLM-driven agent, which simulates local dynamics while selectively referencing others for coordinated reasoning. Three auxiliary agents—Analyzer, Correlator, and Compensator—support summarization, inter-system inference, and residual correction, respectively. This modular, multi-agent design enables interpretable, time-resolved physiological simulation.}
\label{fig:intro}
\end{figure}

As illustrated in Fig.~\ref{fig:intro}, Organ-Agents is built upon a rich collection of heterogeneous, time-resolved intensive care unit (ICU) data derived from 25,064 admissions spanning the Medical Information Mart for Intensive Care (MIMIC)-IV database \cite{johnson2023mimiciv} and two tertiary hospitals, with 125 physiological indicators covering vital signs, lab tests, etc. We systematically model these data into nine organ systems, each encoded by an LLM agent (i.e. Simulator). As the base of Organ-Agents, this design leverages physiological domain knowledge to decouple the high-dimensional time-series clinical data, enabling Simulator agents to evolve on inner-system relationships and adapt to system-specific clinical patterns. 

While modularizing physiological systems simplifies learning, accurate simulation still demands coordination across organs, particularly under shifting clinical states. To support this, we introduce a Correlator agent, which is trained via reinforcement learning to dynamically regulate inter-agent relationships based on patient-oriented context. This mechanism enables each Simulator to selectively access relevant external signals and learn from state-dependent inter-system interactions. Furthermore, to mitigate cumulative simulation errors across long-horizon simulations, we incorporate a Compensator agent, which rectifies the low-confidence simulations with predictive residuals learned from grounded trajectories. This coordination architecture endows Organ-Agents with robust and system-aware simulation capacity. We verify its effectiveness through comprehensive validation across internal held-out patients, external cohorts from distinct clinical settings, and patient groups stratified by severity levels, demonstrating its adaptability and generalization under diverse physiological conditions.

Through structured simulation and dynamic inter-agent coordination, Organ-Agents inherently enforces physiological constraints, enabling the generation of clinically coherent and mechanistically plausible trajectories. Given the complexity of human physiology, such plausibility cannot be captured by a single metric or evaluation perspective. We therefore design a multi-angle evaluation framework that reflects complementary understandings of clinical validity. We first assess the model's ability to reproduce cross-system event chains commonly observed in real-world syndromes, verifying its alignment with objective physiological patterns. This is followed by expert-based assessments, where critical care physicians evaluate the realism and coherence of simulated multi-system trajectories based on clinical experience. Beyond factual plausibility, we simulate counterfactual patient trajectories under hypothetical interventions and examine their internal consistency. Finally, in the context of AI-assisted medicine, we move beyond raw trajectory assessment by evaluating whether simulated data retains plausible predictive features as perceived by machine learning models. Together, these four perspectives, namely (i) clinical patterns, (ii) expert judgment, (iii) counterfactual reasoning, and (iv) AI-based abstraction, jointly substantiate the physiological plausibility and clinical credibility of Organ-Agents’ simulation.

In sum, Organ-Agents offers a feasible path toward simulating dynamic physiological processes through coordinated multi-agent modeling and stepwise temporal reasoning, enabling interpretable, patient-specific inference across interconnected organ systems.

\section{RESULTS}\label{sec2}
Organ-Agents achieved uniformly low simulation errors across nine physiological systems, with the median mean squared error (MSE) below 0.16, and maintained stable performance across all sequential organ failure assessment (SOFA) severity strata. External validation from two tertiary hospitals showed only moderate accuracy drops under distribution shifts. Beyond aggregate metrics, the model accurately reproduced multi-system critical event chains, received high plausibility ratings from critical care experts, and preserved decision-relevant features for downstream AI-based risk simulation.

\subsection{Simulation Accuracy across Multivariate Physiological Indicators}
To evaluate the ability of Organ-Agents to simulate clinically meaningful physiological trajectories, we conducted large-scale simulation experiments across nine physiological systems: respiratory, blood, coagulation, immune, central nervous, cardiovascular, hepatic, renal, and metabolic. Each system is managed by a specialized agent embedded with domain-specific knowledge and adapted to the patient's evolving state. Simulation fidelity was assessed by comparing generated multivariate time-series with ground-truth observations using MSE.

\begin{figure}[h]
\centering
\includegraphics[width=1\linewidth]{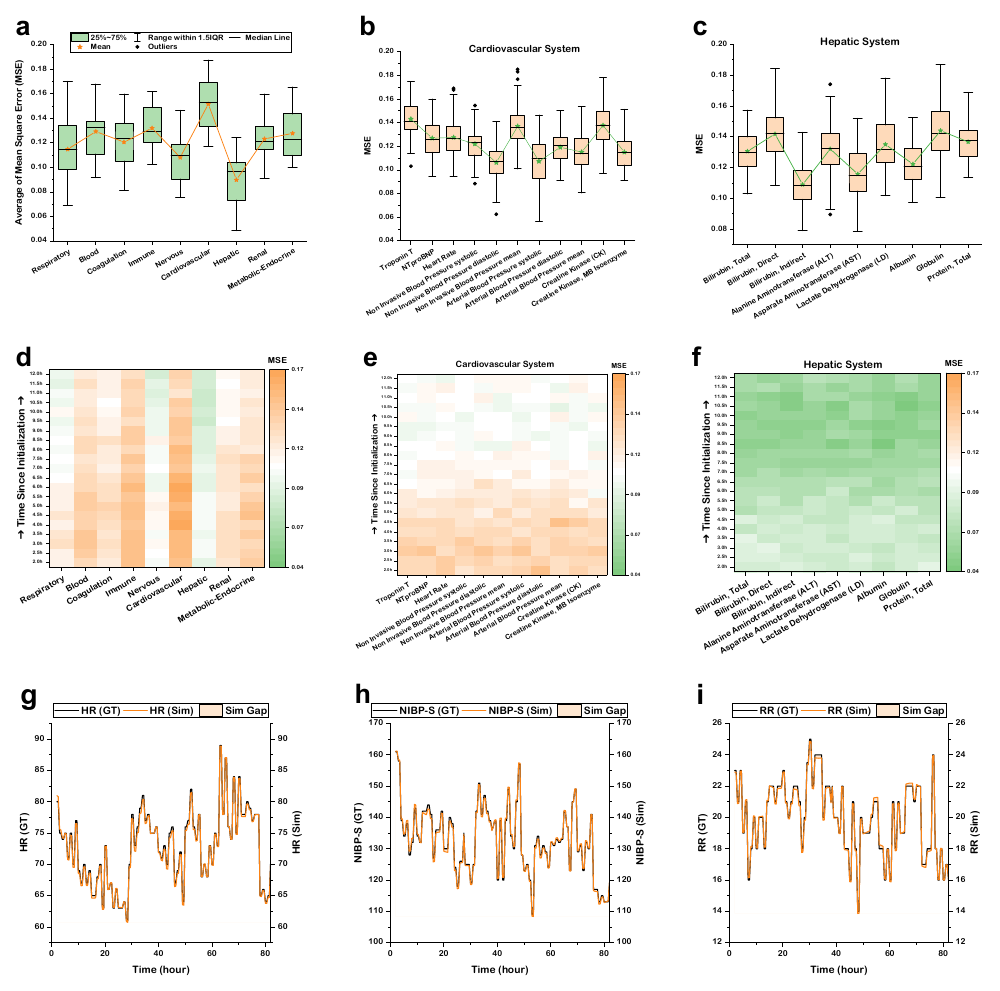}
\caption{Evaluation of Organ-Agents simulation accuracy across physiological indicators. (a) System-level averaged MSE across nine physiological systems. (b,c) Indicator-level MSE in cardiovascular and hepatic systems, respectively. (d) Heatmap of temporal MSE across all systems. (e,f) System-specific MSE heatmaps for cardiovascular and hepatic systems, respectively. (g--i) Simulated vs. real trajectories for heart rate (HR), non invasive blood pressure systolic (NIBP-S), and respiratory rate (RR), respectively. Shaded regions indicate simulation gaps.}
\label{fig:exp1}
\end{figure}

The system-level boxplot, Fig.~\ref{fig:exp1} (a), demonstrates that Organ-Agents maintains uniformly low simulation errors across all systems, with median MSE values consistently below 0.16. Representative system-level indicator distributions, cardiovascular, Fig.~\ref{fig:exp1} (b), and hepatic, Fig.~\ref{fig:exp1} (c), further illustrate the model’s stability across diverse biomarkers. Each system includes a broad array of indicators, and the error distributions suggest effective handling of both frequent and sparse measurements.

Panel (d) presents the overall heatmap of time-evolving MSE for all physiological systems, showing bounded and gradually increasing errors over the 12-hour horizon. Panels (e) through (f) display system-specific heatmaps for the cardiovascular, hepatic, and renal systems, respectively, highlighting intra-system heterogeneity in temporal stability. While most systems maintain low and stable errors, individual indicators may exhibit transient fluctuations, which are effectively constrained by our residual correction agent.

To assess simulation fidelity at the trajectory level, panels (g) through (h) compare simulated versus observed indicator curves for heart rate (HR), non invasive blood pressure systolic (NIBP-S), and respiratory rate (RR). Across all examples, Organ-Agents preserves temporal trends and amplitude ranges, with tight simulation gaps indicating high-resolution tracking. These representative cases underscore the system’s ability to generate physiologically coherent sequences and support traceable, multi-indicator forecasting.

\subsection{Simulation Robustness under Clinical Severity Stratification}
\begin{figure}[htbp]
    \centering
    \includegraphics[page=1,width=\textwidth]{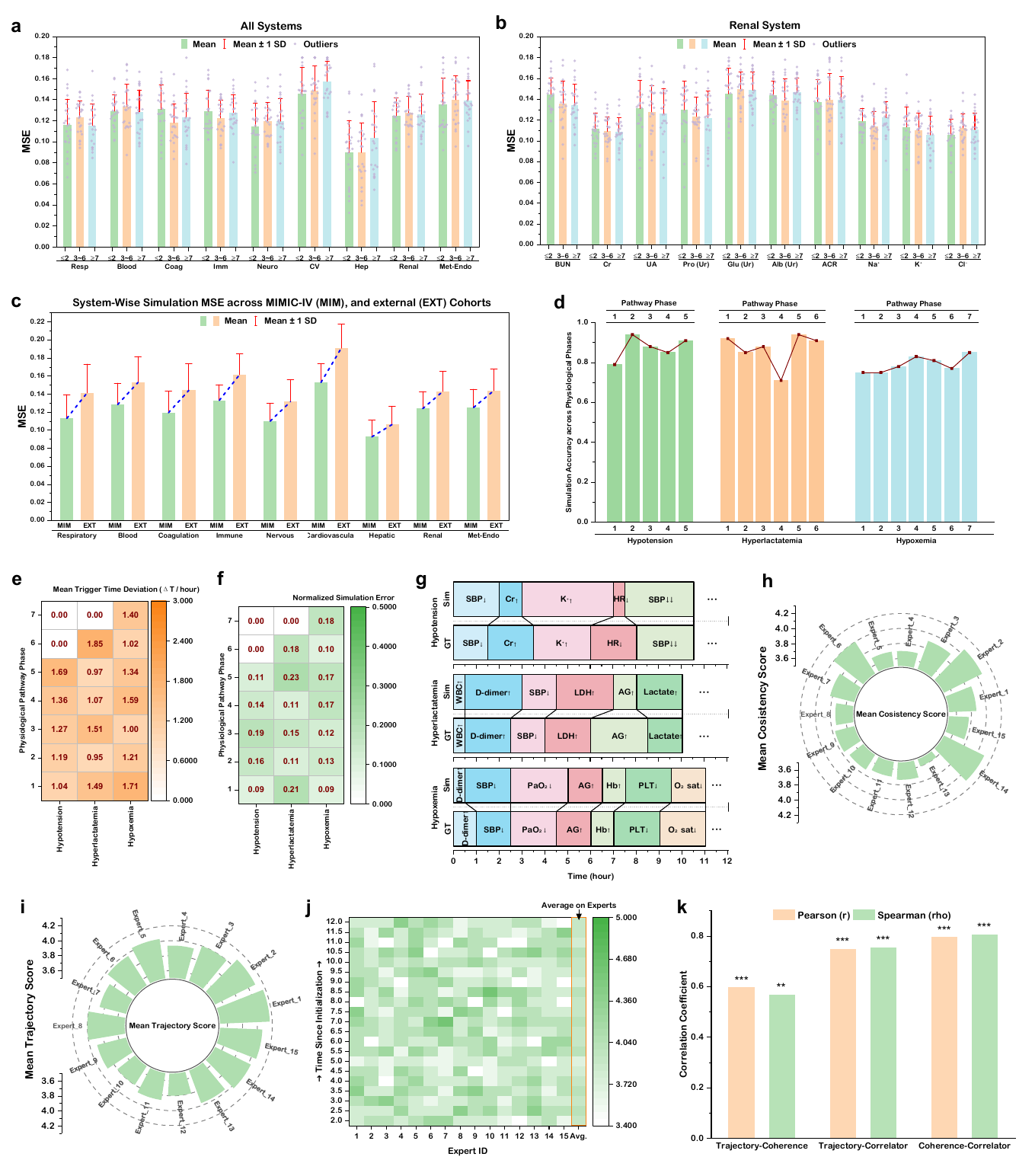}
    \caption{
        \textbf{Simulation results across robustness, trajectory fidelity, and expert evaluation.} 
        (\textbf{a}) Mean simulation error (MSE) across 10 physiological systems under three SOFA severity groups. 
        (\textbf{b}) Simulation error (MSE) for 10 renal indicators across SOFA subgroups. 
        (\textbf{c}) System-wise simulation accuracy (MSE ± SD) of Organ-Agents across internal (MIMIC-IV) and external (TJMUGH, TJCH) cohorts, revealing increased error and variance under cross-center deployment.
        (\textbf{d}) Pathway simulation accuracy for 3 critical physiological trajectories. 
        (\textbf{e}) Trigger time deviation ($\Delta T$) across 3 pathway phases. 
        (\textbf{f}) Normalized simulation error per pathway event. 
        (\textbf{g}) One patient example per pathway: simulated vs. true event sequences. 
        (\textbf{h}) Expert-rated trajectory realism scores. 
        (\textbf{i}) Expert-rated physiological coherence scores. 
        (\textbf{j}) Step-wise reference appropriateness heatmap (experts $*$ steps). 
        (\textbf{k}) Correlation between reference appropriateness, coherence, and trajectory realism.
    }
    \label{fig:exp2}
\end{figure}

To assess the robustness of Organ-Agents across clinically stratified populations, we evaluated its simulation accuracy under varying degrees of disease severity, defined by the SOFA score. Patients were grouped into three strata, SOFA $\leq2$, $3-6$, and $\geq7$, reflecting progressively worsening physiological conditions. As illustrated in Fig.~\ref{fig:exp2} (a) and (b), the model maintained low mean squared error (MSE) across all physiological systems, with only mild increases in error under higher SOFA categories. This trend indicates that Organ-Agents preserves simulation stability even as systemic complexity and volatility increase.

To provide a more granular view, we further examined indicator-level performance within the renal system. The grouped bar plots display the MSE of simulated values for biomarkers such as \textit{urea nitrogen (BUN)}, \textit{creatinine (Cr)}, \textit{albumin, urine (UA)}, and \textit{albumin/creatinine, urine (ACR)} across the same SOFA severity subgroups. Despite slight rises in error at higher severity levels, all indicators remained within a stable error margin, with limited variability as shown by the standard deviation bars. This demonstrates the model’s fine-grained resilience under physiological stress without a breakdown in simulation fidelity.

The visualization reflects consistent performance both at the system-wide and system levels. These results collectively affirm the robustness and adaptive reliability of Organ-Agents across a wide spectrum of clinical conditions, supporting its use in dynamic and high-stakes ICU environments.

\subsection{Cross-Center External Validation of System-Level Simulation Accuracy}
To rigorously assess the generalization capacity of Organ-Agents under realistic deployment conditions, we conducted external validation on the aggregated external cohort, which combines data distributions from two independent hospital datasets—Tianjin Medical University General Hospital and Tianjin Chest Hospital. The same system-level simulation pipeline was applied to both the internal cohort (MIM) and the external cohort (EXT). 

As illustrated in Fig.~\ref{fig:exp2} (c), simulation performance on the EXT cohort exhibits a moderate but systematic increase in mean squared error (MSE) across all physiological systems. The increase is most evident in the cardiovascular, respiratory, and nervous systems, where models tend to exhibit a 20–30\% rise in average MSE compared to internal validation, reflecting higher sensitivity to temporal dynamics and clinical heterogeneity.

Other systems such as immune, coagulation, and renal also show generalization gaps, though with smaller magnitude, suggesting relatively better transferability of simulation logic in those systems. In contrast, hepatic and metabolic-endocrine systems demonstrate the most stable generalization, with minimal MSE drift and largely overlapping error bars, indicating robustness to distribution shifts likely due to lower variability or more standardized clinical protocols.

Importantly, the consistent presence of wider error bars in EXT, particularly in dynamic or sparsely measured systems, underscores the increased uncertainty when models are deployed outside their training distribution. These results suggest that while Organ-Agents retain strong generalization capabilities overall, system-specific variations and real-world data heterogeneity must be explicitly considered to ensure reliable cross-institutional performance.

\subsection{Simulation of Multi-System Critical Event Trajectories} 
To evaluate whether Organ-Agents can faithfully simulate critical physiological event chains, we conducted quantitative experiments across three clinically representative pathways: hypotension progression, hyperlactatemia progression, and hypoxemia progression. Each pathway reflects a multi-system cascade of abnormal indicator transitions, spanning cardiovascular, renal, coagulation, hepatic, metabolic, respiratory, and blood systems. Specifically:

\begin{itemize}
  \item The \textbf{hypotension} pathway involves an initial drop in systolic blood pressure (SBP$\downarrow$, cardiovascular), followed by renal dysfunction indicated by elevated creatinine (Cr$\uparrow$), hyperkalemia (K$^+$$\uparrow$), bradycardia (HR$\downarrow$), and a further decline in blood pressure (SBP$\downarrow\downarrow$).
  
  \item The \textbf{hyperlactatemia} pathway begins with leukocytosis (WBC$\uparrow$, immune), progresses through elevated D-dimer (D-dimer$\uparrow$, coagulation), hypotension (SBP$\downarrow$, cardiovascular), increased lactate dehydrogenase (LDH$\uparrow$, hepatic), widened anion gap (AG$\uparrow$, metabolic), and culminates in elevated serum lactate (Lactate$\uparrow$).
  
  \item The \textbf{hypoxemia} pathway starts with elevated D-dimer (D-dimer$\uparrow$, coagulation) and hypotension (SBP$\downarrow$), followed by reduced arterial oxygen tension (PaO\textsubscript{2}$\downarrow$, respiratory), increased anion gap (AG$\uparrow$), reactive erythrocytosis (Hb$\uparrow$, blood), thrombocytopenia (PLT$\downarrow$), and decreased oxygen saturation (O\textsubscript{2} sat$\downarrow$).
\end{itemize}

For each pathway, patients were first filtered by satisfying the initial trigger condition (e.g., non-invasive systolic blood pressure SBP $<$ 90\,mmHg for hypotension). A 12-hour window of physiological time series was extracted, within which pathway conformity was checked. Patients were included if $\geq$3 events were matched (with tolerance of one-step misordering). Finally, 100 patients were sampled per pathway with diversity in sex, age, and comorbidity.

We adopted three evaluation metrics - \textbf{Pathway Simulation Accuracy}: The proportion of matched events in correct order, with a 3-step grace window. Results are shown in Fig.~\ref{fig:exp2} (d). \textbf{Mean Trigger Time Deviation ($\Delta T$)}: The average latency (in hours) between predicted and observed event onsets (Fig.~\ref{fig:exp2} (e)). \textbf{Normalized Simulation Error}: The average normalized absolute error in indicator value (scaled to physiological range) for each replayed event (Fig.~\ref{fig:exp2} (f)).

Organ-Agents achieved pathway-level mean simulation accuracy of 0.86 (hypotension), 0.79 (hyperlactatemia), and 0.84 (hypoxemia), indicating reliable structural event replay. The mean trigger time deviation remained below 1.9\,h across all phases, and normalized simulation error remained under 0.25, reflecting both timely and physiologically faithful simulations.

Fig.~\ref{fig:exp2} (g) presents one patient per pathway, visualizing simulated vs. ground-truth event trajectories on a unified time axis. For instance, in the hypotension case, Organ-Agents not only correctly predicts the sequence “SBP$\downarrow$ $\rightarrow$ Creatinine$\uparrow$ $\rightarrow$ K$^+\uparrow$ $\rightarrow$ HR$\downarrow$ $\rightarrow$ SBP$\downarrow$$\downarrow$” but also aligns each indicator's temporal dynamics within 0.5–1.5\,h of the true occurrence.

These results validate Organ-Agents's capability in simulating structured multi-organ deterioration processes. It not only simulates the trajectory of critical events, but does so with high signal fidelity and timing precision, providing interpretable simulations for downstream diagnostic and reasoning tasks. 

\subsection{Expert Evaluation of Simulation Credibility}
To assess the clinical plausibility and reasoning interpretability of Organ-Agents's simulated trajectories, we conducted a structured evaluation involving 15 critical care physicians. Each expert independently reviewed 30 patient simulations and provided three categories of 5-point Likert-scale ratings (1 = poor, 5 = excellent): (1) overall trajectory realism, (2) physiological coherence across indicators, and (3) step-wise appropriateness of reference selections by the Correlator agent. Fig.~\ref{fig:exp2} (h)-(k) summarizes the results across all experts and evaluation dimensions.

As shown in the first radial barplot (Fig.~\ref{fig:exp2} (h)), the average trajectory realism scores were consistently concentrated between 3.8 and 4.2. This suggests that most of the simulated indicator sequences were judged clinically plausible as standalone trajectories. Minor inter-expert variation was observed, reflecting individual thresholds for trajectory acceptability.

The second radial plot (Fig.~\ref{fig:exp2} (i)) illustrates system-level physiological coherence ratings. While the overall trend remained positive (mean scores between 3.7 and 4.2), the variance across experts was slightly higher than in the trajectory ratings, indicating differing perceptions of internal consistency among co-evolving indicators. Experts emphasized that certain organ systems (e.g., renal or hepatic) required more stringent co-regulation to be perceived as physiologically sound.

To evaluate whether the Correlator agent’s reference selections aligned with clinical intuition, each expert reviewed 24 step-wise decisions on 5 held-out patients. The resulting expert-step matrix (Fig.~\ref{fig:exp2} (j)) shows high per-step credibility, with most reference scores clustered between 3 and 5. The rightmost column aggregates average reference quality across all experts, further supporting the consistency of reference logic.

We further investigated the relationship between reference appropriateness and simulation quality using both Pearson and Spearman correlations (Fig.~\ref{fig:exp2} (k)). Notably, reference scores were strongly correlated with both trajectory realism (Pearson $r=0.749$, Spearman $\rho=0.756$) and physiological coherence (Pearson $r=0.796$, Spearman $\rho=0.807$), all with $p<0.001$. This finding supports the hypothesis that high-quality reference selection contributes directly to generating simulations that are both clinically plausible and physiologically consistent. Moderate correlation was also observed between trajectory and coherence scores ($r=0.598, p<0.001; \rho=0.568, p<0.01$), suggesting that realism and coherence are partially independent yet mutually reinforcing dimensions.

\subsection{Evaluation on Counterfactual Simulation}

To evaluate the counterfactual simulation capability of the model, we take the fluid resuscitation strategies for sepsis patients (early fluid resuscitation vs. delayed fluid resuscitation) as an example, comparing the physiological system indicators and mortality risk differences between simulated and real patient environments, exploring the simulation accuracy and reliability of the simulator under different strategies.

First, sepsis patients are divided into two groups based on the actual fluid resuscitation initiation time recorded in the MIMIC IV database: the early fluid resuscitation group (ER group, patients who received fluid resuscitation within 1 hour of sepsis diagnosis) and the delayed fluid resuscitation group (DR group, patients who received fluid resuscitation 3 to 6 hours after diagnosis). Then, using the model, we generate predicted survival trajectories for each patient in an alternative scenario. Specifically, for patients in the ER group, we simulate their physiological response if they had received delayed resuscitation. For patients in the DR group, we simulate their response under early fluid resuscitation. Next, using propensity score matching based on patient characteristics, we match simulated ER group patients with real DR group patients, and simulated DR group patients with real ER group patients. We then perform a comparative analysis of the mean squared error of physiological system indicators and APACHE II scores \cite{knaus1985apache} between the simulated and real patient groups.

\begin{figure}[htbp]
\centering
\includegraphics[width=1\textwidth]{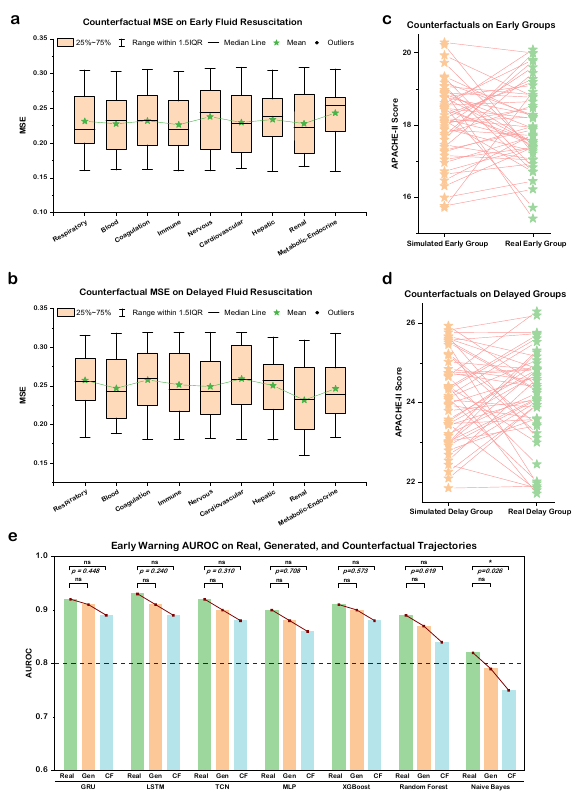}
\caption{
\textbf{Evaluation of simulator-based early fluid resuscitation recommendations and clinical semantic alignment with early warning models.}
(a) Mean square error of physiological system indicators between the simulated early group and the actual early group.
(b) Mean square error of physiological system indicators between the simulated delayed group and the actual delayed group.
(c) APACHE II scores between the simulated early group and the actual early group.
(d) APACHE II scores between the simulated delayed group and the actual delayed group.
(e) AUROC scores of AI-based early warning models evaluated across real, generated, and counterfactual patient states, showing high consistency across domains.
}
\label{fig:exp3}
\end{figure}

Fig.~\ref{fig:exp3} (a) and (b) present the mean squared errors of physiological system indicators for the simulated early group versus the real early group, and the simulated delayed group versus the real delayed group, respectively. The average error ranges from 0.16 to 0.31 and from 0.18 to 0.32, indicating a small difference between the simulation results and the real data. Fig.~\ref{fig:exp3} (c) and (d) show the differences in APACHE II scores between the simulated early group and the real early group, and between the simulated delayed group and the real delayed group, respectively. Most of the connecting lines are nearly horizontal, indicating that the differences in APACHE II scores between the two groups are not significant. These results demonstrate that the simulator has strong counterfactual simulation capabilities.

\subsection{Semantic Fidelity of Generated Trajectories under Early Warning Models}
To evaluate whether the simulated trajectories generated by Organ-Agents preserve clinically meaningful semantics from the perspective of downstream AI-based early warning systems, we assessed the transferability of real-trained classifiers to synthetic and counterfactual domains. This serves as an indirect validation of the physiological plausibility of generated data—specifically, whether they retain class-discriminative patterns similar to those in real-world trajectories.

To ensure consistent labeling across all domains, we uniformly re-applied the Sepsis-3 diagnostic criteria to both generated and counterfactual sequences, even though true outcomes for counterfactuals are inherently unobservable.

As shown in Fig.~\ref{fig:exp3} (e), seven representative classifiers—Gated Recurrent Unit (GRU) \cite{cho2014gru}, Long Short-Term Memory (LSTM) \cite{hochreiter1997long}, Temporal Convolutional Network (TCN) \cite{bai2018empirical}, Multilayer Perceptron (MLP) \cite{rumelhart1986mlp}, XGBoost \cite{chen2016xgboost}, Random Forest \cite{breiman2001random}, and Naive Bayes \cite{mccallum1998comparison}—were trained on real-world trajectories and evaluated on generated (Gen) and counterfactual (CF) data. Most models demonstrated only marginal degradation in Area Under the ROC Curve (AUROC) across domains, suggesting strong semantic alignment. For instance, GRU achieved AUROCs of 0.92, 0.91, and 0.89 on Real, Gen, and CF data, respectively; LSTM and TCN showed similar patterns with CF drops within 0.04. Even shallow models such as MLP and Random Forest preserved adequate performance (e.g., MLP: 0.90 $\rightarrow$ 0.86). In contrast, Naive Bayes exhibited the most pronounced performance decline (0.82 $\rightarrow$ 0.75), with a statistically significant difference between Real and CF ($p = 0.026$), indicating its heightened sensitivity to distributional shifts.

These results indicate that the synthetic trajectories generated by Organ-Agents preserve sufficient semantic structure to support stable classifier performance under domain shift. Our objective is not to validate the factual correctness of each counterfactual outcome, but to assess whether simulated trajectories—when labeled using shared clinical criteria—exhibit semantic fidelity to real-world data distributions. The overall minimal degradation and consistent transferability across most models suggest no significant semantic drift, confirming that Organ-Agents generates trajectories that remain interpretable and clinically coherent from the perspective of the downstream AI applications.

\section{DISCUSSION}

Organ-Agents was developed to address a fundamental gap in clinical Artificial Intelligence: the ability to simulate complex, multi-system physiological trajectories in a transparent and controllable manner. Rather than framing medicine as a sequence of isolated tasks—such as diagnosis, summarization, or question answering—Organ-Agents aims to model human physiology as an evolving system of interdependent agents. By integrating agentic language models with graph-based structural priors and hierarchical memory, the system offers a unified platform for forward simulation, contrastive reasoning, and trajectory traceability.

Recent years have witnessed rapid expansion of large language models (LLMs) into clinical domains, yet current applications remain fragmented. In diagnostic support, Med-PaLM and its successors have achieved near-expert performance on benchmark datasets \cite{singhal2023large}, while GPT-4 variants have shown competitive accuracy in differential diagnosis across real-world vignettes \cite{goh2024large}. Summarization-focused studies, such as those by Hom et al., demonstrate LLMs outperforming clinicians in note condensation \cite{van2023clinical}, and models like ChatGPT have been preferred over physicians in patient-facing QA due to perceived empathy \cite{ayers2023comparing}. In more specialized domains, multimodal LLMs such as Med-Flamingo \cite{moor2023med} and LLaVA-Med \cite{li2023llava} have shown promise in fusing imaging and textual data. Yet across these efforts, models operate largely as static text processors. They lack a consistent internal representation of physiological dynamics, offer limited support for longitudinal reasoning, and rarely incorporate explicit structural priors. Importantly, they do not simulate – they predict. No prior work has systematically addressed the challenge of modeling temporally evolving, multi-system clinical states with interpretability and modularity. Organ-Agents thus fills a critical void by aligning LLM capacity with physiological abstraction, offering a scalable foundation for patient-centric simulation and explanation.

\subsection{Research Impacts}
Organ-Agents introduces three key innovations. First, its agentic architecture mirrors human physiology through discrete yet coordinated LLM agents, each responsible for a specific system. Second, by encoding system dependencies in an individual agent, it ensures coherence across organ-level dynamics. Third, through counterfactual sampling based on similarity-matched cases, the model supports contrastive exploration of disease evolution. These design choices yield a system capable of not only predicting plausible future states, but also tracing the reasoning paths that lead to them.

We evaluated Organ-Agents’s simulation stability by comparing its performance across clinical severity strata defined by SOFA scores. As shown in Fig. \ref{fig:exp2} (a), the model maintained consistently low simulation error even in high-acuity states ($\mathrm{SOFA} \geq 7$), demonstrating resilience to physiological volatility. Organ-Agents also preserved cross-system consistency, demonstrating coherent physiological responses across organ systems without localized degradation. These findings affirm that embedding physiological priors into language-based simulation meaningfully enhances robustness under complex conditions.

In addition to forward simulation, Organ-Agents was assessed on its capacity to reconstruct clinical event chains. As illustrated in Fig. \ref{fig:exp2} (f), for canonical trajectories such as sepsis-induced hypotension or hypoxia-linked lactic acidosis, the model accurately recovered temporal marker progressions. Moreover, by generating matched counterfactual trajectories from a library of similar patients, it enabled clinicians to explore how alternative baseline conditions could have influenced the evolution of system states and provided a tractable form of counterfactual traceability. This capacity to compare ``what happened" versus ``what could have happened" supports clinical hypothesis generation, decision validation, and educational introspection. 

We further investigated the credibility of Organ-Agents through qualitative expert review and structured output analysis. Clinicians rated the model’s outputs highly on trajectory realism, and physiological coherence, as shown in Fig. \ref{fig:exp2} (g)-(i). The multi-agent design allowed individual systems to generate localized rationales, while central agents (e.g., Analyzer, Correlator) synthesized global summaries and inter-system linkages. This modularity facilitated intuitive inspection of physiological transitions, such as linking hypovolemia to renal perfusion drop. Nevertheless, we observed cases where transitions between states lacked granularity or context, suggesting a need for more fine-tuned temporal anchoring and prompt alignment. 

The strength of Organ-Agents lies not merely in LLM scale, but in the structure-function synergy it achieves. Its architecture promotes modular reasoning, physiological consistency, and temporal continuity, qualities absent in most existing clinical LLMs. The inclusion of a RL-controlled interaction ensures inter-agent coordination, while layered memory on the prompt structure enables sustained state awareness. Unlike end-to-end black-box models, Organ-Agents permits direct attribution of simulation outputs to system-level mechanisms, offering a viable path toward AI systems clinicians can trust and audit.

\subsection{Clinical Impacts}
The development of Organ-Agents represents a significant advancement in the field of clinical medicine, offering a powerful tool for understanding and predicting the complex dynamics of human physiology. By simulating the intricate interplay of multiple organ systems, Organ-Agents provides clinicians with a digital twin platform that can enhance diagnostic accuracy, support personalized treatment planning, and facilitate proactive patient management.

One of the most profound clinical impacts of Organ-Agents lies in its ability to enable early detection of life-threatening conditions. Through high-fidelity simulation of physiological trajectories, clinicians can identify subtle patterns and deviations that may precede critical events such as sepsis-induced hypotension or hypoxemia-linked lactic acidosis. This capability is particularly valuable in ICU settings where early recognition of physiological derangements can significantly alter clinical outcomes. The model's validated accuracy in reproducing complex event chains provides clinicians with actionable insights that conventional monitoring systems often fail to deliver until later stages.

Moreover, Organ-Agents serves as a valuable tool for evaluating the effects of different treatment strategies. By generating counterfactual patient trajectories, clinicians can explore how alternative treatments might have influenced disease progression. This is especially relevant for time-sensitive interventions like sepsis resuscitation, where the simulator's ability to maintain physiological coherence across organ systems offers a unique advantage for therapeutic decision-making. The model's preservation of clinical semantic alignment further ensures that simulated treatment responses remain clinically plausible.

The model's capacity for longitudinal simulation also aids in understanding disease evolution over time. This can lead to a deeper understanding of disease mechanisms and the development of more effective treatment protocols. Additionally, the interpretability of Organ-Agents's simulations fosters trust among clinicians, making it more likely to be adopted in real-world clinical settings. The strong agreement between simulated trajectories and expert clinical judgment demonstrates the model's potential as both an educational tool and clinical decision aid.

In summary, Organ-Agents has the potential to transform critical care by providing clinicians with a sophisticated, interpretable, and adaptable tool for physiological modeling. By enabling real-time simulation of multi-organ interactions, it offers unprecedented opportunities for personalized intervention planning and dynamic treatment optimization in complex critical illness. Its impact extends beyond individual patient care to advancing medical research and education, ultimately contributing to the broader goal of precision medicine.

\subsection{Limitations}
%Several limitations remain. All evaluations were conducted on retrospective, de-identified electrical health records (EHR) data, which limits insight into real-time clinical applicability and patient-specific feedback integration. The model currently assumes static system configurations, posing challenges for adaptation to rare diseases or atypical multi-organ presentations. Its architecture is not yet optimized for high-dimensional multimodal inputs such as imaging or waveforms.

%Future research will explore integrating Organ-Agents with dynamic sensory data and real-time clinical feedback \cite{rosenthal2025rethinking}, enabling a tighter simulation-decision loop. Enhancing adaptability through physician-in-the-loop reinforcement \cite{feng2025doctoragent}, extending to rare system combinations \cite{mehandru2024evaluating}, and incorporating structured uncertainty modeling \cite{liu2025uncertainty, savage2025large} are key priorities. We also envision benchmarking Organ-Agents within standardized clinical simulation challenges that emphasize trajectory fidelity and causal traceability \cite{gallifant2025tripod}. By advancing both mechanistic reasoning and generative adaptability, Organ-Agents aspires to become a robust foundation for interpretable, adaptive, and system-aware clinical AI.  Moreover, this framework may enable the discovery of previously unrecognized inter-system associations during disease onset, facilitating novel insights into disease pathophysiology and system-level interactions.

Several limitations remain. First, all current evaluations were conducted retrospectively using de-identified electronic health record (EHR) data, which constrains assessment of the simulator’s real-time clinical utility and its ability to incorporate patient-specific feedback dynamically. Second, the model assumes a static multi-agent system architecture, limiting its generalizability to rare diseases, atypical organ interactions, or evolving pathophysiological states. Furthermore, the current implementation is not yet optimized for high-dimensional, multimodal inputs such as medical imaging, biosignals, or unstructured clinical notes.

To address these limitations, future work will focus on integrating dynamic sensory inputs and real-time clinical feedback into the Organ-Agent framework, thereby enabling a tighter simulation-decision loop \cite{rosenthal2025rethinking}. Key directions include incorporating physician-in-the-loop reinforcement learning \cite{feng2025doctoragent}, extending the framework to model rare and complex system-level presentations \cite{mehandru2024evaluating}, and embedding structured uncertainty modeling to enhance safety and robustness \cite{liu2025uncertainty, savage2025large}. Additionally, we aim to benchmark simulator performance in standardized clinical simulation tasks that prioritize trajectory fidelity, intervention interpretability, and causal traceability \cite{gallifant2025tripod}.

Critically, we are planning a prospective randomized controlled trial (RCT) to evaluate the simulator’s performance in real clinical settings. In this study, patients will be randomly assigned to either a “simulator-assisted care” group—where physicians are supported by the simulator in decision-making—or a “standard care” group relying solely on conventional clinical judgment. Outcome measures will include length of stay and survival rate. We hypothesize that simulator-assisted care will lead to improved patient outcomes, particularly in terms of survival and organ function restoration.

In the long term, this framework may uncover novel inter-system relationships underlying complex diseases, offering fresh insights into disease mechanisms and enhancing system-aware clinical reasoning. By advancing both mechanistic fidelity and adaptive generalization, Organ-Agents aspires to serve as a robust foundation for interpretable, real-time, and personalized clinical AI.

\section{METHODS}\label{sec11}
To construct a robust multi-agent physiological simulation system leveraging large language models (LLMs), we propose a systematic methodological framework integrating expert clinical knowledge, longitudinal clinical data, and sophisticated predictive modeling. This framework is specifically tailored to simulate complex physiological and pathological dynamics across multiple interconnected human systems. 

\subsection{Data Source and Study Population}

\subsubsection{Data Source}
The data used in this study was obtained from the Medical Information Mart for Intensive Care IV (MIMIC-IV, version 2.2), a publicly available, de-identified clinical database \cite{johnson2023mimic}. MIMIC-IV contains clinical information for patients admitted to the intensive care unit (ICU) at Beth Israel Deaconess Medical Center (BIDMC) in Boston, USA, between 2008 and 2019, encompassing a total of 180,733 unique hospital admissions \cite{johnson2023mimiciv}. The database is structured with a modular design, consisting of five modules and a total of 96 tables. It includes a wide range of variables such as patient demographics, laboratory test results, medication records, vital signs, surgical procedures, disease diagnoses, medication management, and survival status. Our institution waived the requirement for BIDMC Institutional Review Board (IRB) approval and informed consent, as investigators who meet the data use requirements are granted access to MIMIC-IV under existing IRB approval (Researcher Certification Code: 59450069). Additionally, external validation was performed using a multi-center dataset obtained from Tianjin University Chest Hospital and the General Hospital Affiliated with Tianjin Medical University. The study complied with the principles of the Declaration of Helsinki (2013 revision) and was approved by the Ethics Committees of both Tianjin University Chest Hospital and Tianjin Medical University General Hospital (Approval No. 2020YS-022-01). Given the retrospective and observational nature of the study, the requirement for informed patient consent was waived.

\subsubsection{Study Population}
\begin{figure}[h]
    \centering
    \includegraphics[width=1\linewidth]{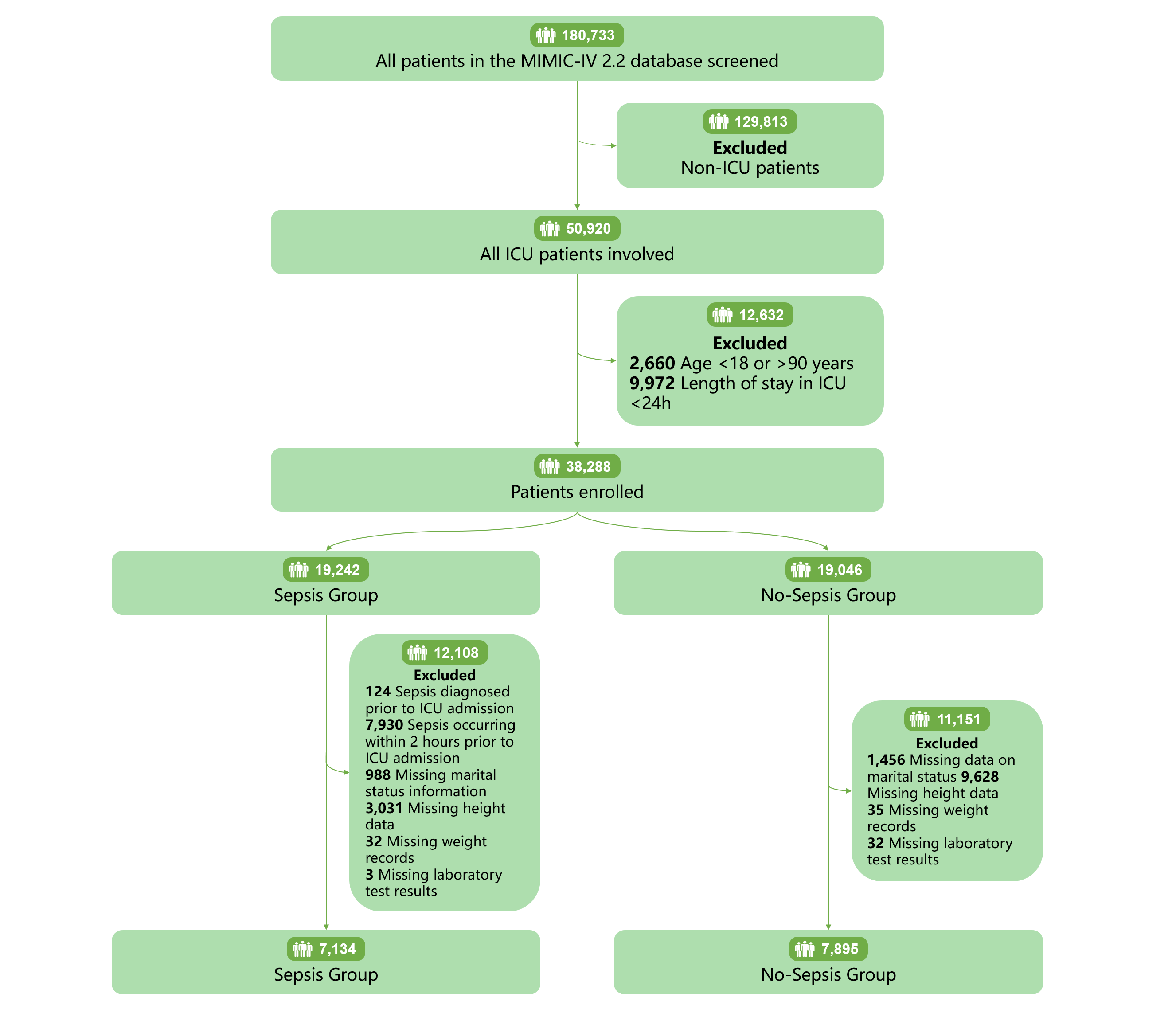}
    \caption{Flow chart of the exclusion criteria applied to the MIMIC-IV patient cohort. Out of 180,733 initially screened patients, 38,288 eligible ICU patients were identified after applying key exclusion criteria based on age, ICU stay duration, and data completeness. These patients were then divided into Sepsis (n=7,134) and No-Sepsis (n=7,895) groups, with further exclusions based on sepsis onset timing and missing clinical or demographic data. Sepsis classification was based on the Sepsis-III definition.}
    \label{flowchart}
\end{figure}

The flowchart of exclusion criteria in the MIMIC-IV cohort is depicted in Fig. \ref{flowchart}. Initially, 180,733 patients were screened. After excluding 129,813 non-ICU patients, 50,920 ICU patients remained for further analysis. Subsequently, 2,660 patients aged \textless 18 or \textgreater 90 years, as well as 9,972 patients with an ICU length of stay of less than 24 hours, were excluded, resulting in 38,288 eligible patients. These patients were then categorized into two groups: the Sepsis Group (n=19,242) and the No-Sepsis Group (n=19,046). Sepsis was defined based on the Sepsis-III criteria  \cite{singer2016sepsis3}, which includes suspected infection accompanied by a sequential organ failure assessment (SOFA) score $\geq2$ points. Within the Sepsis Group, additional exclusions were applied: 124 patients with pre-ICU sepsis diagnoses, 7,930 patients with sepsis onset within 2 hours before ICU admission, and 988, 3,031, 32, and 32 patients missing data on marital status records, height data, weight data, or laboratory test results, respectively. This resulted in a final Sepsis Group of 7,134 patients. In the No-Sepsis Group, exclusions included 1,456 patients missing marital status records, 9,628 missing height data, 35 missing weight data, and 33 with missing laboratory test results, yielding a final cohort of 7,895 patients.

The figure presents the inclusion and exclusion criteria flowchart for the external validation dataset. A total of 45,232 patients admitted to ICU at Tianjin Chest Hospital and the Affiliated General Hospital of Tianjin Medical University between January 1, 2021, and December 31, 2023, were initially screened. Of these, 22,543 patients were excluded for not meeting the inclusion criteria. The primary exclusion reasons were as follows: (1) age \textless 18 years (n=1,534); (2) a diagnosis of sepsis prior to ICU admission (n=891); (3) discharge or death within 24 hours of admission due to causes unrelated to sepsis (n=1,096); (4) inability to obtain baseline measurements (n=2,912); (5) missing time-series indicators during the ICU stay (n=11,853); (6) lack of information regarding sepsis diagnosis or mortality outcomes (n=4,257). Ultimately, 22,689 patients met the eligibility criteria and were included in the study. Based on the occurrence of sepsis during their ICU stay, participants were categorized into a sepsis group (n=10,736) and a non-sepsis group (n=11,953).

\subsubsection{Data collection}
The data were extracted from the MIMIC-IV database using the Structured Query Language (SQL) with the PostgreSQL database management system. The following parameters were systematically collected for all patients: (1) Demographic characteristics, including age, gender, race, insurance status, marital status, height, weight, body mass index (BMI), body surface area (BSA), smoking history, and alcohol consumption; (2) Comorbidities and complications, including obesity, hypertension, diabetes, hyperlipidemia, hypothyroidism, stroke, heart failure, myocardial infarction, peripheral vascular disease, chronic obstructive pulmonary disease (COPD), acute renal failure, and tumors; (3) Laboratory test results, including complete blood count, blood biochemistry, arterial blood gas analysis, coagulation function tests, liver and kidney function tests, urinalysis, and results of microbial cultures of blood samples; (4) Vital signs data, including heart rate, systolic blood pressure, diastolic blood pressure, mean arterial pressure, non-invasive systolic blood pressure, non-invasive diastolic blood pressure, non-invasive mean pressure, respiratory rate, body temperature, and Glasgow Coma Scale (GCS) score; (5) Treatment medication status, including common antibiotics, hemodynamic drugs, anticoagulants, and sedatives; (6) Outcome parameters, including ICU admission time, ICU discharge time, sequential organ failure assessment (SOFA) score, and time of sepsis onset. Among these, laboratory test results, vital signs data, and medication use were dynamic time-series sequences.

The external validation dataset includes the following 63 variables: (1) General information, including gender, age, height, weight, BMI, educational level, occupation, marital status, smoking history, alcohol consumption, Montreal Cognitive Assessment (MoCA) score, Mini-Mental State Examination (MMSE) score, metabolic equivalent, and the European System for Cardiac Operative Risk Evaluation score (EuroSCORE); (2) Comorbidities, including obesity, diabetes, hypertension, myocardial infarction, arrhythmia, coronary artery disease, stroke, COPD, chronic kidney disease, thyroid dysfunction, peripheral vascular disease, gastrointestinal disorders, and cancer; (3) Laboratory tests, including pH, pCO2, pO2, K+, HCO3-, blood glucose, subacute bacterial endocarditis (SBE), hemoglobin, platelet count, hematocrit, alanine transaminase, aspartate transaminase, albumin, blood urea nitrogen, creatinine, uric acid, creatine kinase, lactate dehydrogenase, hydroxybutyrate dehydrogenase, cardiac troponin T, C-reactive protein, and B-type natriuretic peptide; (4) Imaging studies, including Left Ventricular Ejection Fraction (LVEF), aortic sinus anteroposterior diameter, left main coronary artery, and left anterior descending artery; (5) Treatment medications, including commonly used antibiotics, hemodynamic drugs, anticoagulants, and sedatives; (6) Vital signs data, including cardiac output, cardiac index, stroke volume, stroke index, systemic vascular resistance, systemic vascular resistance index, systolic blood pressure, diastolic blood pressure, mean arterial pressure, pulse pressure variability, heart rate, heart rate variability, cerebral oxygen saturation, and tissue oxygen saturation; (7) Outcome measures, including incidence of sepsis and mortality. Among these, vital signs data were dynamic time-series sequences.

\subsubsection{Data preprocessing}
Data preprocessing in this study was performed using Python. Data preprocessing primarily consists of two components: handling static variables and processing time series data. For static variables, missing values were removed to ensure data integrity. Binary variables (such as gender, smoking, alcohol consumption, hypertension, diabetes, etc.) were encoded as binary numbers (0 and 1). Multi-class variables (such as race, marital status, and ICU location) were converted into machine-readable format using one-hot encoding. For continuous variables such as height and weight, outliers were treated as missing values, and multiple imputation methods were used to handle missing values. Additionally, new features (e.g., BMI and BSA) are derived from existing variables to enrich the dataset and incorporate physiologically meaningful indicators.

Extract all time series data for all patients from admission to the ICU to discharge from the ICU (or onset of sepsis), primarily including laboratory test results, vital signs data, and treatment drug data. Based on the SOFA scoring principles and clinical physician experience, all extracted time-series variables were categorized into nine major systems: respiratory system, bloodal system, coagulation system, immune system, nervous system, cardiovascular system, hepatic system, renal system, and metabolic and endocrine system. The specific variables for each system are detailed in Table \ref{tab: variables}. Due to inconsistent sampling frequencies across different data sources, all time series variables were resampled at a fixed 30-minute interval to achieve time alignment. Given the inherent sparsity of clinical time series data, forward imputation was used to fill in missing values. Since the impact of missing values may increase over time, this study employed masked decay to reduce the influence of missing values, applying stronger decay to missing values farther from the nearest valid data point for each patient. After preprocessing, the complete dataset was randomly partitioned into training, validation, and test sets using a 7:1.5:1.5 ratio.

\begin{table}[h]
    \centering
    \caption{The specific variables for each system.}
    \label{tab: variables}
    \begin{tabular}{l l p{8cm}} % 8cm width for the Variables column
        \toprule
        \textbf{Category} & \textbf{Number} & \textbf{Variables} \\ 
        \midrule
        Respiratory system & 6 & pH, pCO2, pO2, Calculated Total CO2, Respiratory Rate, O2 saturation pulseoxymetry \\ 
        Blood system & 7 & Red Blood Cells, Hemoglobin, Hematocrit, RDW, MCV, MCH, MCHC \\ 
        Coagulation system & 10 & Platelet Count, Lactate, Thrombin, PT, Fibrinogen, Functional, PTT, PT(INR), D-Dimer, Enoxaparin (Lovenox), Heparin Sodium (Prophylaxis) \\ 
        Immune system & 24 & White Blood Cells, Absolute Neutrophil Count, Absolute Lymphocyte Count, Monocytes, Eosinophils, Basophils, Vancomycin, Ampicillin/Sulbactam, Azithromycin, Aztreonam, Cefazolin, Cefepime, Ceftazidime, Ceftriaxone, Ciprofloxacin, Clindamycin, Gentamicin, Levofloxacin, Linezolid, Meropenem, Piperacillin, Piperacillin/Tazobactam, Tobramycin, Voriconazole \\ 
        Nervous system & 8 & Temperature Fahrenheit, GCS, Endotracheal intubation, Midazolam, Fentanyl, Hydromorphone, Propofol, Morphine Sulfate \\ 
        Cardiovascular system & 17 & Heart Rate, Non Invasive Blood Pressure systolic, Non Invasive Blood Pressure diastolic, Non Invasive Blood Pressure mean, Arterial Blood Pressure systolic, Arterial Blood Pressure diastolic, Arterial Blood Pressure mean, Creatine Kinase (CK), Creatine Kinase MB Isoenzyme, Troponin T, NTproBNP, Epinephrine, Dobutamine, Dopamine, Phenylephrine, Norepinephrine, Vasopressin \\ 
        Hepatic system & 9 & Bilirubin (Total), Bilirubin (Direct), Bilirubin (Indirect), Alanine Aminotransferase, Aspartate Aminotransferase, Lactate Dehydrogenase, Albumin, Globulin, Protein (Total) \\ 
        Renal system & 10 & Urea Nitrogen, Creatinine, Uric Acid, Protein, Glucose Urine, Albumin Urine, Albumin/Creatinine Urine, Sodium, Potassium, Chloride \\ 
        Metabolism and endocrine system & 9 & Glucose Blood, HbA1c, Triglycerides, Cholesterol (Total), Cholesterol-HDL, Cholesterol-LDL, Total Calcium, Free Calcium, Anion Gap \\ 
        \bottomrule
    \end{tabular}
\end{table}

\subsubsection{Statistical analysis}

Statistical analyses were performed using SPSS 24.0 software. Continuous variables are presented as medians with interquartile ranges, and group comparisons were made using the Mann-Whitney U test. Categorical variables are expressed as frequencies and percentages, with comparisons between groups performed using the chi-square test or Fisher's exact test. All statistical tests were two-sided, and a \textit{p}-value of less than 0.05 was considered statistically significant. 

\subsection{Organ-Agents}

\begin{figure}[t]
    \centering
    \includegraphics[width=1\linewidth]{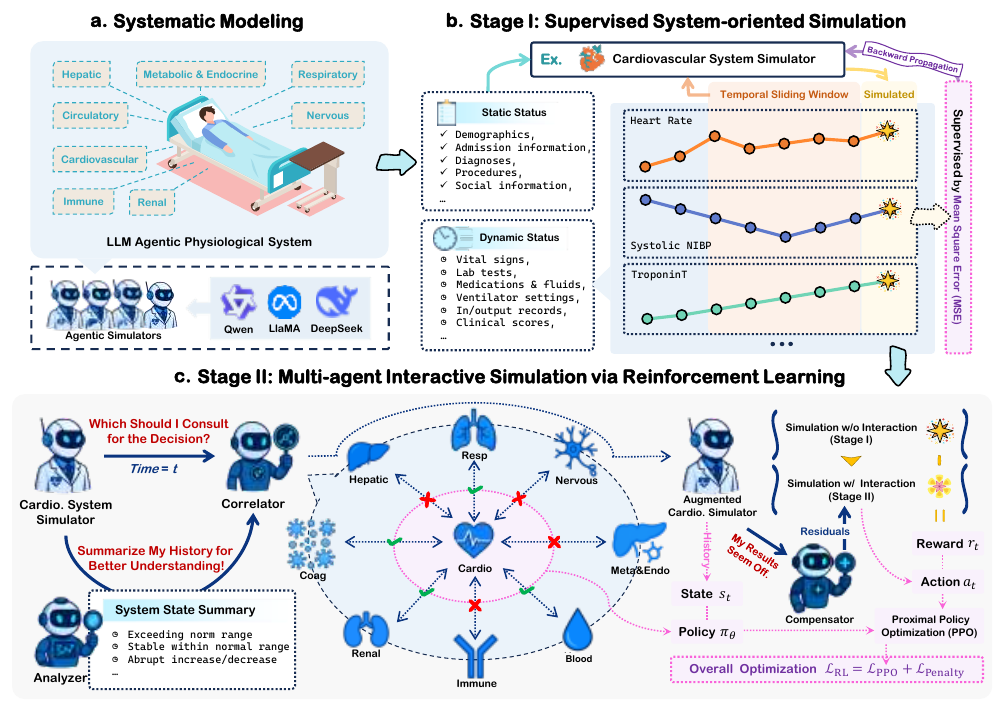}
    \caption{Overview of the Organ-Agents framework. Stage I performs system-specific supervised simulation using individual LLM agents trained on structured static and dynamic features. Stage II introduces multi-agent interaction via reinforcement learning, where modules such as the Analyzer, Correlator, and Compensator facilitate inter-system reasoning and residual refinement. The simulation proceeds in a temporally recursive manner, optimizing the policy via proximal policy optimization (PPO) to generate physiologically coherent and clinically plausible trajectories.}
    \label{framework}
\end{figure}

We propose \textbf{Organ-Agents}, a multi-agent large language model framework for simulating physiological system dynamics through stepwise, interpretable simulations. Each organ system is represented by an autonomous LLM agent that performs structure-aware forecasting, selective inter-agent referencing, and residual error correction (Fig.~\ref{framework} (a)). Training proceeds in two stages: (1) supervised fine-tuning (SFT) for single-system baseline simulation (Fig.~\ref{framework} (b)); and (2) multi-agent interactive simulation via reinforcement learning (Fig.~\ref{framework} (c)). This architecture mirrors the modular, interdependent nature of human physiology—each system processes its own indicators while selectively incorporating information from others to capture complex cross-organ interactions. Unlike monolithic models, Organ-Agents supports modular retraining, enhanced interpretability, and scalable integration of physiological systems.

In the SFT stage, each agent is instantiated as a decoder-only LLM (e.g., Qwen3 \cite{qwen3}) and fine-tuned via low-rank adaptation (LoRA) \cite{hu2022lora} using system-specific time-series data. The agent generates next-step simulations for its corresponding indicators, along with a self-estimated confidence score, jointly emitted in a single output sequence. 

While local dynamics are often sufficient for short-term forecasting, many physiological indicators are shaped by inter-system dependencies (e.g., mean arterial pressure reflects both cardiac output and renal regulation). To capture these effects, the second stage introduces a reinforcement-learned policy agent, Correlator, that enables agents to autonomously select and incorporate external variables when beneficial. Considering that the usefulness of external references is both time-dependent and individual-specific, requiring reasoning over the entire temporal span, we introduce a system status Analyzer that condenses historical trends and salient clinical events into concise, structured summaries. This abstraction reduces input redundancy and enhances the model’s ability to interpret contextual signals. Furthermore, as stepwise simulation is prone to early-stage simulation errors that propagate over time, we deploy a residual correction agent, Compensator, to dynamically adjust low-confidence outputs. This agent refines simulations based on contextual cues, serving as an adaptive mechanism to curb cumulative error. 

\subsubsection{Stage I: Supervised System-oriented Simulation}

Although physiological signals exhibit strong inter-system coupling, much of the local dynamics, such as circadian fluctuations, autoregulation, or system-specific intervention, can be effectively learned in isolation. Meanwhile, the progressive and structured nature of time-series data poses unique challenges for LLMs, which are not inherently designed for temporal simulation or strict format adherence. Therefore, the first stage of Organ-Agents is designed to help each system agent internalize the underlying temporal patterns and simulation rules of its own physiological indicators in a standalone setting. 

\begin{figure}[t]
    \centering
    \includegraphics[width=1\linewidth]{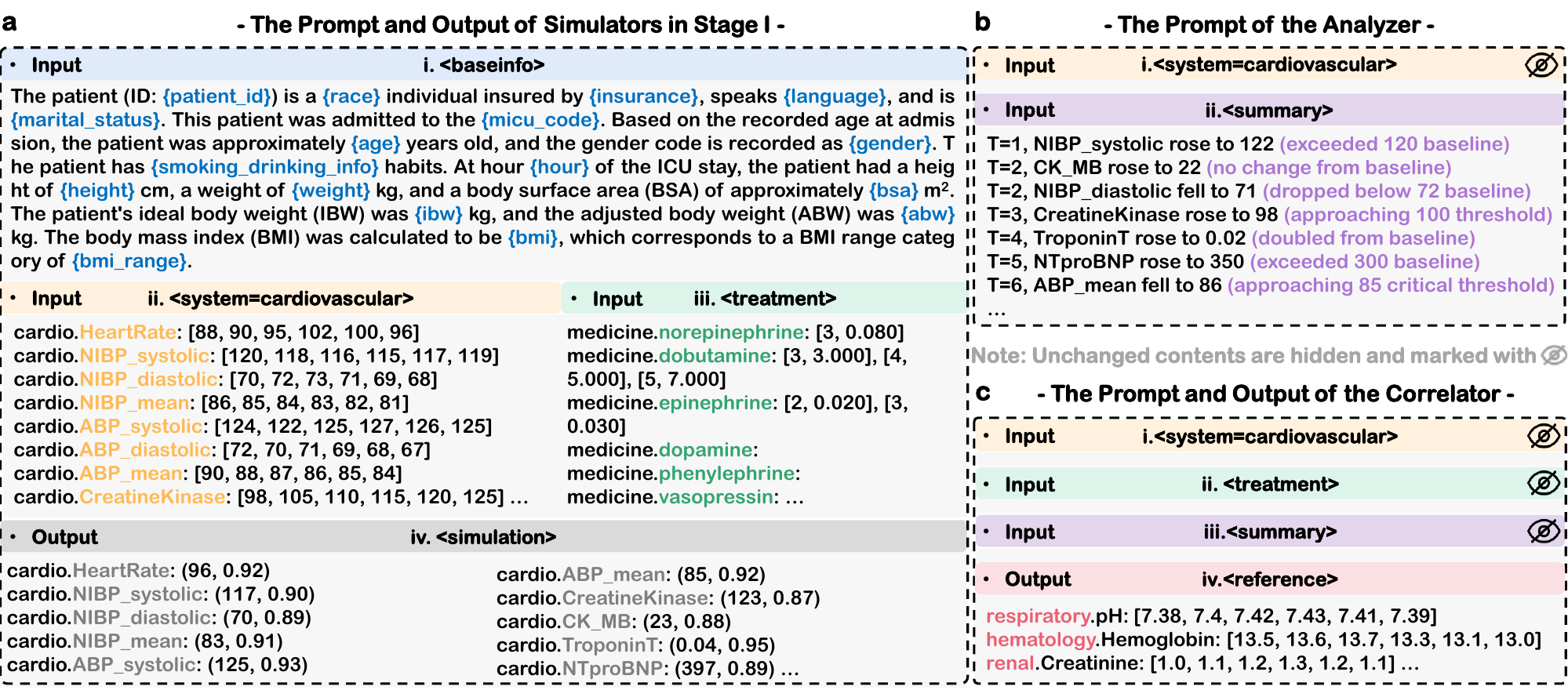}
    \caption{Prompts and output formats of Simulator (Stage I), Analyzer and Correlator. 
    (a) The Simulator receives structured patient information, including demographic profile (\texttt{<baseinfo>}), time-series status (\texttt{<system=cardiovascular>}), and treatment records (\texttt{<treatment>}), and produces simulations for key indicators along with confidence scores (\texttt{<simulation>}).
    (b) The Analyzer summarizes clinically significant changes in indicator trajectories, identifying abnormal events based on predefined thresholds (\texttt{<summary>}).
    (c) The Correlator generates supplementary prompts that incorporate cross-system signals (e.g., from respiratory, blood, and renal systems; \texttt{<reference>}) to support downstream analysis.}
    \label{StageIwAgents}
\end{figure}

To help each system agent internalize temporal patterns and learn to comply with structured simulation rules, we fine-tune the LLM with low-rank adaptation (LoRA) to perform single-step simulation based on domain-specific time-series signals.  Specifically, low-rank trainable adapters (rank = 4) are inserted into the attention and multi-layer perceptron (MLP) modules of the base LLM. This allows us to update fewer than 1\% of parameters while preserving the core capabilities of the pre-trained backbone. LoRA fine-tuning is applied jointly across all system agents using a shared prompt format, facilitating scalable and modular deployment. As shown in Fig. \ref{StageIwAgents} (a), each input prompt is carefully designed to encode both static and dynamic contextual information, enabling the model to reason over short-term trends while grounding simulations in relevant physiological priors. 

The prompt begins with a \texttt{<baseinfo>} block, which encodes static patient attributes such as age, sex, race, BMI, and relevant medical history. This metadata provides the model with individualized physiological priors, enabling it to contextualize time-series signals within patient-specific baselines. By conditioning on these attributes during SFT, the model learns to simulate system dynamics across diverse populations while avoiding implausible extrapolations. The structured design of the \texttt{<baseinfo>} block also promotes consistent prompt formatting and facilitates generalization to previously unseen demographic combinations.

Following this, a \texttt{<system=X>} block presents a sliding window of recent time-series values for all indicators within a specific organ system. This segment captures local physiological dynamics, including circadian fluctuations, short-term autoregulation, and transient responses to interventions. Treatment records are appended in a dedicated \texttt{<treatment>} block, with medication names and timestamps aligned to the indicator window. This structure allows the model to learn how system-specific interventions modulate future trajectories under varying contexts.

The overall training objective in Stage~I comprises two components. First, we use the standard token-level cross-entropy loss
\[
\mathcal{L}_{\mathrm{CE}} = - \sum_{t=1}^T \log p_\theta(y_t \mid x, y_{<t}),
\]
which ensures that the generated output sequence matches the target tokens, enforcing both lexical correctness and adherence to the structured $(\hat{y}_t, \hat{c}_t)$ output format.  
Second, to adapt the model to the numerical nature of physiological simulation, we introduce a \textit{constraint loss} $\mathcal{L}_{\mathrm{cons}}$ consisting of two terms:  
(a) a numerical constraint encouraging accurate simulations of indicator values via $\mathrm{MSE}(\hat{y}_t, y_t)$; and  
(b) a confidence calibration constraint aligning the predicted confidence $\hat{c}_t$ with the inverse-error target $\tilde{c}_t = \exp(-|\hat{y}_t - y_t|)$:
\[
\mathcal{L}_{\mathrm{cons}} = \mathrm{MSE}(\hat{y}_t, y_t) + \lambda \cdot \mathrm{MSE}\left( \hat{c}_t, \tilde{c}_t \right),
\]
where $\lambda$ balances predictive accuracy and confidence calibration.

The complete Stage~I loss is thus
\[
\mathcal{L}_{\mathrm{SFT}} = \mathcal{L}_{\mathrm{CE}} + \mathcal{L}_{\mathrm{cons}},
\]
which enables each system agent to produce syntactically valid and numerically accurate forecasts while self-assessing their reliability, laying the foundation for selective communication and error-aware coordination in subsequent multi-agent interaction.

\subsubsection{Stage II: Multi-agent Interactive Simulation via Reinforcement Learning}

After Stage I SFT, Organ-Agents acquires a basic understanding of the internal dynamics of each organ system. However, this stage remains a mechanistic imitation: the agents operate in isolation, without modeling inter-system dependencies. As a result, submodels remain overly compartmentalized, and the framework lacks the ability to explain cross-system causal pathways—for example, how declining renal function may elevate blood pressure due to fluid retention and sympathetic activation. In reality, inter-system interactions are complex, time-varying, and not easily reducible to fixed medical rules. Therefore, in Stage II, we leverage the flexibility of LLMs to let Organ-Agents autonomously discover dynamic inter-system referencing patterns, allowing it to construct globally coherent, patient-specific simulations.

Concretely, we aim for each organ simulator to dynamically decide which other systems and indicators to reference at every timestep, based on its own history and evolving physiological state. To this end, we introduce a reinforcement learning–driven multi-agent coordination mechanism. Building upon independently trained system Simulators, we add three auxiliary agents, Analyzer, Correlator, and residual Compensator, to support memory consolidation, cross-system reference selection, and adaptive error compensation, respectively. 

Starting from the initial state, the Analyzer monitors all system indicators and summarizes notable trends when significant shifts occur. These summaries are automatically recorded and later retrieved as historical prompts. While the Analyzer primarily relies on the LLM’s intrinsic capacity for temporal summarization, we perform a small-scale supervised fine-tuning to enable consistent, schema-compliant formatting of its outputs. This light adaptation requires minimal training cost and preserves the general summarization ability of the underlying model. An example prompt and output are shown in Fig. \ref{StageIwAgents} (b):

During simulation, each organ simulator's prompt is extended with selected external indicator histories to support cross-system coordination. The candidate indicators and their originating systems are selected by the Correlator, a policy agent trained via reinforcement learning using proximal policy optimization (PPO) \cite{schulman2017ppo}. At each timestep $t$, the Correlator receives the structured prompt (see an example in Fig. \ref{StageIwAgents} (c)) encoding the decision state $s_t$, which includes three components:
\begin{itemize}
\item \textbf{System State} (\texttt{<system=X>}): a sliding window of recent local indicator values specific to the target organ system;
\item \textbf{Historical Summary} (\texttt{<summary>}): event-level summaries generated by the Analyzer, capturing recent notable changes;
\item \textbf{Medical Intervention} (\texttt{<treatment>}): timestamps and categories of administered drugs, representing recent therapeutic actions that may affect system dynamics.
\end{itemize}

\begin{figure}[h]
    \centering
    \includegraphics[width=1\linewidth]{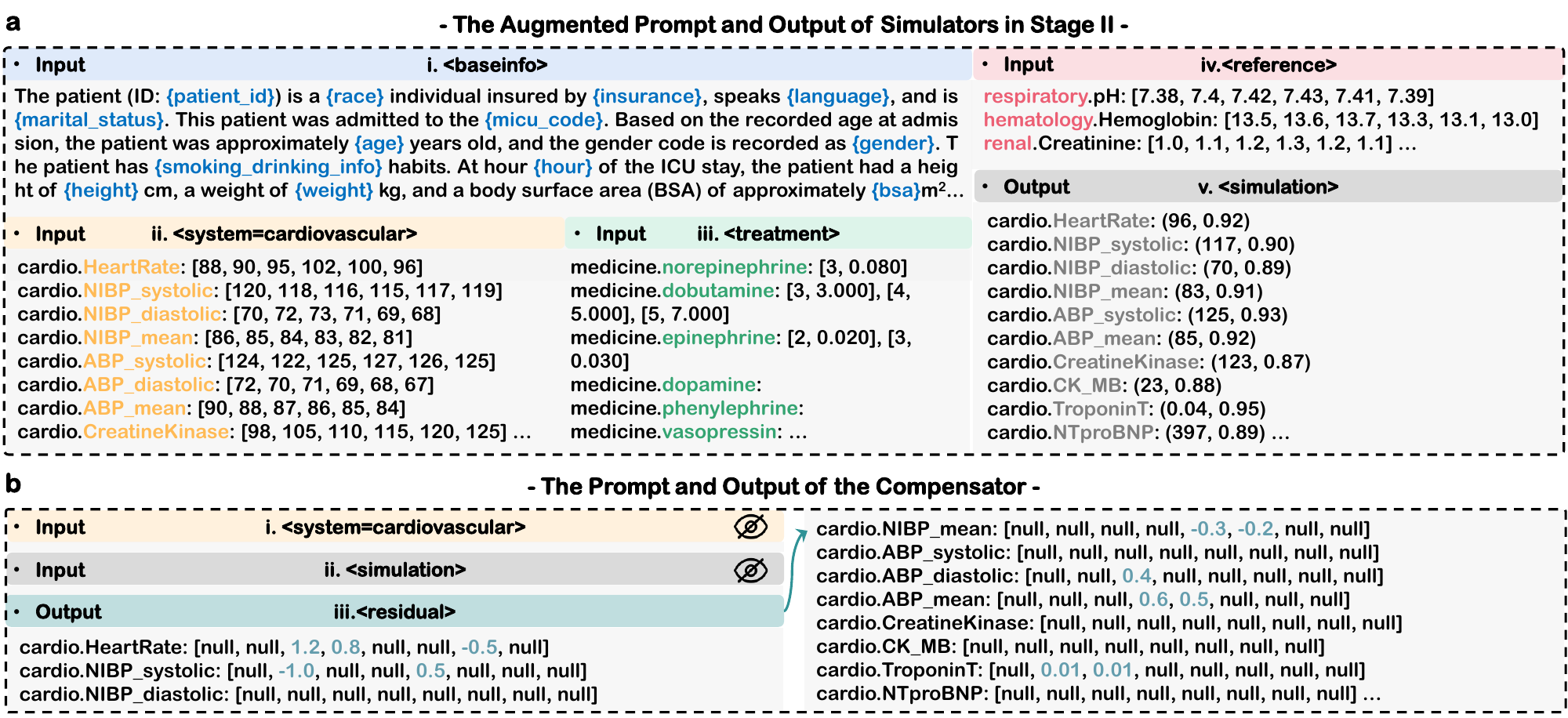}
    \caption{Enhanced simulation process in Stage II, incorporating multi-agent referencing and post-simulation refinement.
    (a) The Simulator receives an augmented prompt composed of patient base information, cardiovascular measurements, treatments, and actively integrated cross-system references (e.g., pH, hemoglobin, creatinine). It produces updated simulations with associated confidence scores.
    (b) The Compensator estimates residual errors for each indicator by comparing simulated simulations with expected dynamics, as reflected in the \texttt{<residual>} field. The residual values are used to refine selected low-confidence outputs (e.g., NIBP\_mean, ABP\_mean, TroponinT), resulting in compensated simulations that better align with inter-system trends.}
    \label{StageII}
\end{figure}

These components enable the Correlator to evaluate both internal dynamics and external contextual signals. Based on the resulting state representation $s_t$, the policy network outputs a probability distribution over the candidate reference indicator set $\mathcal{V}_t$, from which a subset is sampled to form the action $a_t$. The selected indicators $a_t$ are then injected into the system Simulator’s prompt to enable agentic interaction across physiological systems, as illustrated in Fig. \ref{StageII} (a).

Based on the augmented prompt, the system Simulator generates a new simulation $\hat{y}_t$. In parallel, a baseline simulation $\hat{y}_t^{(0)}$, computed from Stage I (i.e. without cross-system references), is produced. The reward is then defined as the reduction in simulation error:
\begin{equation}
r_t = \mathrm{MSE}(\hat{y}_t^{(0)}, y_t) - \mathrm{MSE}(\hat{y}_t, y_t).
\end{equation}
This reward encourages the Correlator to select reference indicators that meaningfully improve predictive accuracy.

However, although this reward formulation minimizes immediate simulation error, even small deviations in earlier timesteps can accumulate over time, resulting in compounding errors that are difficult to correct in later stages of simulation. To address this issue, we introduce a residual compensation agent, Compensator, that estimates per-step simulation errors and adjusts simulations accordingly.

To ensure computational efficiency, we use the system Simulator's confidence scores to determine whether residual Compensator is necessary. When triggered, the residual Compensator receives a structured prompt that includes the system Simulator's current simulation, recent indicator history, and a log of past compensation attempts,  as illustrated in Fig. \ref{StageII} (b). Based on this input, it outputs an estimate of the expected simulation error $\hat{e}_t$.

The Correlator's parameters are optimized using PPO algorithm. Let $\pi_{\theta}$ denote the current policy and $\pi_{\theta_{\text{old}}}$ the policy used to generate the action $a_t$. The clipped surrogate objective is defined as:
\begin{equation}
\mathcal{L}_{\mathrm{PPO}} = - \mathbb{E}_t \left[ \min \left( \rho_t A_t, \text{clip}(\rho_t, 1 - \epsilon, 1 + \epsilon) A_t \right) \right],
\quad \rho_t = \frac{\pi_\theta(a_t|s_t)}{\pi_{\theta_{\text{old}}}(a_t|s_t)},
\quad A_t = r_t - b_t.
\end{equation}
Here, $\epsilon$ is the clipping range that limits the change in policy probability ratios between updates to stabilize training, set to $0.2$ in our experiments following common PPO practice. The term $b_t$ is a running baseline estimate used to reduce the variance of the policy gradient. In our implementation, it is computed as an exponential moving average (EMA) of past rewards:
\begin{equation}
b_t = \alpha \cdot r_{t-1} + (1 - \alpha) \cdot b_{t-1},
\end{equation}
where $\alpha \in (0,1)$ is a smoothing coefficient, set to $0.9$ in our experiments to balance responsiveness to recent rewards with stability against noise. This provides a reliable approximation of the expected reward without the additional complexity of training a separate value function, yielding an advantage estimate $A_t$ that quantifies how much better the selected references perform compared to average outcomes.

To further encourage exploration and promote sparse referencing, we regularize the policy using two additional terms: an entropy bonus and an $L_1$-norm sparsity penalty. The entropy term, $\mathcal{H}[\pi(a_t|s_t)]$, encourages the policy to maintain uncertainty over reference selections, thereby mitigating premature convergence and fostering a broader search of the action space. Conversely, the sparsity term $\|a_t\|_1$ penalizes excessive referencing, promoting minimal and focused cross-system interactions to improve both computational efficiency and interpretability.

The final reinforcement learning objective for the Correlator is:
\begin{equation}
\mathcal{L}_{\mathrm{RL}} = \mathcal{L}_{\mathrm{PPO}} + \beta_{\text{sparsity}} \cdot \|a_t\|_1 - \beta_{\text{entropy}} \cdot \mathcal{H}[\pi(a_t|s_t)],
\end{equation}
where $\beta_{\text{sparsity}} = 0.015$ and $\beta_{\text{entropy}} = 0.005$ are empirically chosen to balance parsimony and exploration. This setting yields stable training while encouraging selective yet sufficiently diverse referencing patterns that reflect physiological dependencies.

To avoid training interference, the residual Compensator is optimized independently using a standard mean squared error objective. Importantly, the Compensator operates as a post-hoc correction module and is optimized independently from the main system Simulator. As such, it does not influence the gradient flow of the primary simulating model during backpropagation. Its goal is to estimate and reduce the simulation error between the system Simulator's simulation $\hat{y}_t$ and the ground truth $y_t$. The loss is defined as:
\begin{equation}
\mathcal{L}_{\mathrm{Res}} = \left( \hat{e}_t - \mathrm{MSE}(\hat{y}_t, y_t) \right)^2.
\end{equation}
This helps prevent error propagation during long-term simulation and ensures temporal stability across multi-step forecasting trajectories. To obtain the final corrected simulation, the output of the Compensator is added to the original system Simulator’s simulation. Specifically, the compensated simulation result $\hat{y}_t^{\text{sim}}$ is computed as:
\begin{equation}
\hat{y}_t^{\text{sim}} = \hat{y}_t + \hat{e}_t,
\end{equation}
where $\hat{y}_t$ is the system Simulator's direct simulation, and $\hat{e}_t$ is the Compensator’s estimated residual. This additive correction adjusts for bias and accumulated errors from earlier simulation steps, thereby improving both accuracy and stability over long time horizons.

Algorithm \ref{alg:stage2} summarizes the complete pipeline of Stage II. It integrates dynamic reference selection by the Correlator, policy optimization via PPO, and confidence-gated residual correction from the Compensator, forming a unified loop for coordinated multi-agent simulation.

\begin{algorithm}[h!]
\caption{Reinforcement-Guided Multi-agent Interaction.}
\label{alg:stage2}
\KwIn{Patient's basic information $B$, System state history $H_t$, Analyzer log $L_t$, Treatment records $T_t$, System simulation $\hat{y}_t$ and confidence  $\hat{c}_t$}
\KwOut{Compensated simulation $\hat{y}_t^{\text{sim}}$}
\BlankLine
\tcp{Correlator Inference}
% \Indp
    Construct state $s_t$ from $\left(H_{t-1}, L_{t-1}, T_t\right)$ \;
    Sample reference action $a_t \sim \pi_\theta(a_t | s_t)$ \;
    Inject reference indicators into Simulator: $\left(B, H_{t-1}, T_t, a_t\right)$\;
    Obtain updated simulation $\hat{y}_t$ using augmented input\;
% \Indm
\BlankLine
\tcp{Reward Computation and Policy Update}
% \Indp
    Generate baseline simulation $\hat{y}_t^{(0)}$ using local input only\;
    Compute reward: $r_t = \mathrm{MSE}(\hat{y}_t^{(0)}, y_t) - \mathrm{MSE}(\hat{y}_t, y_t)$ \;
    Update baseline estimate: $b_t = \alpha \cdot r_{t-1} + (1 - \alpha) \cdot b_{t-1}$\;
    Compute advantage: $A_t = r_t - b_t$ \;
    Update $\pi_\theta$: $\mathcal{L}_{\mathrm{PPO}} = - \mathbb{E}_t \left[ \min \left( \rho_t A_t, \text{clip}(\rho_t, 1 - \epsilon, 1 + \epsilon) A_t \right) \right]$\;
    Regularization: $\mathcal{L}_{\mathrm{RL}} = \mathcal{L}_{\mathrm{PPO}} + \beta_{\text{sparsity}} \cdot \|a_t\|_1 - \beta_{\text{entropy}} \cdot \mathcal{H}[\pi(a_t|s_t)]$\;
% \Indm
\BlankLine
\tcp{Residual compensation (Conditional)}
\If{$\hat{c}_t$ $<$ confidence threshold}{
    Construct residual Compensator prior using $\left(H_{t-1}, \hat{y}_t, \hat{e}_{<t}\right)$\;
    Predict error: $\hat{e}_t$ = Compensator$(\cdot)$\;
    Compute loss: $\mathcal{L}_{\text{Res}} = \left( \hat{e}_t - \mathrm{MSE}(\hat{y}_t, y_t) \right)^2$ \;
    Update residual Compensator parameters via gradient descent\;
    Adjust simulation: $\hat{y}_t^{\text{sim}} = \hat{y}_t - \hat{e}_t$
}
\Else{
    Set $\hat{y}_t^{\text{sim}} = \hat{y}_t$
}
\Return{$\hat{y}_t^{\text{sim}}$}
\end{algorithm}

\subsection{Implementation Details}

\subsubsection{Agent Core Selection}
The selection of the agent core is pivotal to the overall performance of the agent system in our Organ-Agents framework. LLMs serving as agent cores must meet several critical requirements to function effectively: (1) robust multi-turn interactive dialogue capabilities to support real-time decision integration and refinement; (2) substantial domain-specific knowledge, supported by a parameter scale sufficient to encompass comprehensive medical understanding for informed decision-making; and (3) a context window of at least 10k tokens to accommodate the extensive tool descriptions within our framework, as models with shorter contexts have proven inadequate for this purpose. In addition, open-source availability, tokenizer robustness, and compatibility with parameter-efficient adaptation methods such as LoRA are essential for practical deployment across diverse clinical environments.

These criteria form the basis for our subsequent evaluation of candidate LLMs, covering general-purpose, domain-specific medical, and proprietary models. The evaluation results, presented in Appendix~\ref{sec:llm_eval}, confirm that \textbf{Qwen3-8B}\cite{qwen3} offers the most balanced trade-off between instruction alignment and physiological fidelity. As the latest installment in the Qwen series, Qwen3-8B supports a 128k-token context window, demonstrates strong multilingual instruction-following and long-context reasoning capabilities, and is fully compatible with LoRA-based adaptation. Its open-source nature and robust generalization performance make it an ideal backbone for constructing modular and interpretable agent-based systems in Organ-Agents.

\subsubsection{Evaluation Metrics}
To rigorously assess the performance of Organ-Agents across various dimensions of physiological simulation, we employed a suite of evaluation metrics that are detailed below. These metrics were specifically chosen to align with the clinical and technical objectives of our study, ensuring a comprehensive and objective evaluation of the model's capabilities.

\vspace{0.15cm}\textbf{Simulation Accuracy Metrics.} 
The fidelity of Organ-Agents's simulations was primarily evaluated using the mean squared error (MSE) \cite{che2018recurrent,rajkomar2018scalable} between the model's simulation and the observed values across all physiological indicators. This metric quantifies the average squared difference between the simulated and actual trajectories, providing a direct measure of the model's precision. Organ-Agents achieved a system-wide average MSE of 0.12 across multivariate time series, indicating a high degree of fidelity in replicating real patient trajectories. The MSE was also computed for each physiological system individually, with median values consistently below 0.16 across all systems.

\vspace{0.15cm}\textbf{Temporal Stability Metrics.}
Inspired by previous work \cite{futoma2017improved}, we adopt temporal MSE heatmaps to evaluate the model's performance over time. These visualizations provided insights into how the simulation errors evolved over the 12-hour horizon, revealing bounded and gradually increasing errors that highlight the model's temporal stability.

\vspace{0.15cm}\textbf{Clinical Severity Stratification Metrics.}
The robustness of Organ-Agents under varying degrees of disease severity was assessed using the sequential organ failure assessment (SOFA) score. Patients were stratified into three groups based on their SOFA scores ($\leq2$, $3-6$, and $\geq7$), and the model's simulation accuracy was evaluated within each group. The results demonstrate that Organ-Agents maintains stable performance across all severity levels, with only mild increases in error under higher SOFA categories.

\vspace{0.15cm}\textbf{Critical Event Trajectory Simulation Metrics.}
Drawing inspiration from the evaluation metrics in \cite{murtaza2023synthetic}, Organ-Agent's ability to simulate multi-system critical event trajectories was evaluated using three key metrics:
\begin{itemize}
\item Pathway Simulation Accuracy: This metric measures the proportion of correctly sequenced events within a 3-step grace window. Organ-Agents achieved pathway-level simulation accuracies of 0.86 for hypotension, 0.79 for hyperlactatemia, and 0.84 for hypoxemia.
\item Mean Trigger Time Deviation: This metric quantifies the average latency (in hours) between predicted and observed event onsets. The model maintained a mean trigger time deviation below 1.9 hours across all phases.
\item Normalized Simulation Error: This metric assesses the average normalized absolute error in indicator values for each replayed event, scaled to the physiological range. The error remained under 0.25, reflecting both timely and physiologically faithful simulations.
\end{itemize}

\vspace{0.15cm}\textbf{Expert Evaluation Metrics.}
As emphasized by \cite{liu2020reporting}, the clinical plausibility and interpretability of the simulated trajectories were evaluated by 15 critical care physicians using a 5-point Likert scale (1 = poor, 5 = excellent). The evaluation focused on three dimensions:
\begin{itemize}
\item Overall Trajectory Realism: Experts rated the realism of the simulated indicator sequences. The average score was consistently above 3.8 across all experts.
\item Physiological Coherence: Experts assessed the coherence of co-evolving indicators across systems. The average scores ranged between 3.7 and 4.2.
\item Reference Appropriateness: The step-wise reference selections by the Correlator agent were evaluated. Most reference scores were clustered between 3 and 5, indicating high credibility.
\end{itemize}

\vspace{0.15cm}\textbf{Clinical Semantic Alignment Metrics.}
The semantic alignment \cite{esteban2017real} of Organ-Agents-generated patient trajectories with real-world distributions was evaluated using AUROC scores of seven classifiers trained on real data and tested on generated and counterfactual data. The results showed minimal performance degradation across domains, which indicates that the model preserves sufficient class-discriminative structure to support downstream classification tasks.

\subsubsection{Training Details}
All experiments were conducted on a computing platform equipped with an NVIDIA H20 GPU (96 GB VRAM), running Ubuntu 20.04 and CUDA 12.4. The implementation is based on Python 3.12, utilizing PyTorch 2.5.1 and Huggingface Transformers 4.50. Parameter-efficient tuning is enabled via the LoRA technique, implemented through the PEFT library \cite{peft}. 

The first stage of training involves SFT of Qwen-3-8B, on system-specific clinical simulation tasks. The input data consist of time-windowed clinical records stored in JSONL format. Each training sample comprises a prompt, which includes static patient information and treatment history, and a response representing the simulation output for a designated physiological system. Samples not involving the target system are excluded to ensure focused learning. LoRA adapters are added to the backbone model, and the tokenizer is initialized accordingly. The training procedure is implemented using a modified Huggingface Trainer, which combines the standard supervised cross-entropy loss with an additional supervision loss to extract structured values and confidence levels from the generated text. The optimizer is AdamW, with a learning rate of $2\times10^{-5}$, a warm-up ratio of 0.03, and a maximum gradient norm of 1.0. Training is performed for 3 epochs with a batch size of 4, and FP16 mixed-precision training is enabled to improve memory efficiency. 

The second training stage adopts a reinforcement learning paradigm based on PPO, aiming to further align the model’s behavior with downstream objectives. During PPO training, the data are grouped by patient and time window to maintain temporal coherence. Each batch includes input prompts, ground truth outputs, system tags, patient identifiers, and time indices. Three Qwen3-8B based agents, Analyzer, Correlator, and Compensator are initialized with Qwen-3-8B, and then are fine-tuned with LoRA during the second training stage. A customized trainer class governs the optimization loop, which proceeds as follows: first, the raw input IDs are decoded to reconstruct the original prompt. This prompt is augmented by inserting a high-level summary generated by an Analyzer agent, followed by a candidate block listing non-target variables. The modified prompt is passed to the Correlator agent which identifies relevant variables that are then matched against historical records to generate a structured reference block. 

This reference-enhanced prompt is evaluated by the reward model (typically the Simulator agent), which assigns a scalar reward to each sample based on the alignment between generated output and clinical expectations. The policy model generates responses conditioned on the same input. The PPO loss is formulated as the sum of a clipped policy loss, a sparsity regularization term and an entropy regularization term. Invalid or ill-formed samples (e.g., those lacking valid references) are filtered out. PPO training is conducted using a batch size of 8, with learning rates of $10^{-5}$ for the policy network. The clip range is set to 0.2, and the maximum sequence length is 1024 tokens, with generated outputs capped at 256 new tokens.

% \section{Conclusion}\label{sec13}
% \backmatter

%\section*{Declarations}

%Some journals require declarations to be submitted in a standardised format. Please check the Instructions for Authors of the journal to which you are submitting to see if you need to complete this section. If yes, your manuscript must contain the following sections under the heading `Declarations':

%\begin{itemize}
%\item Conflict of interest/Competing interests (check journal-%specific guidelines for which heading to use)
%\item Ethics approval and consent to participate
%\item Consent for publication
%\item Data availability 
%\end{itemize}


\begin{appendices}

\section{Evaluation of Pretrained LLMs for Organ-Agents} \label{sec:llm_eval}

\begin{figure}[htbp]
    \centering
    \includegraphics[width=1\linewidth]{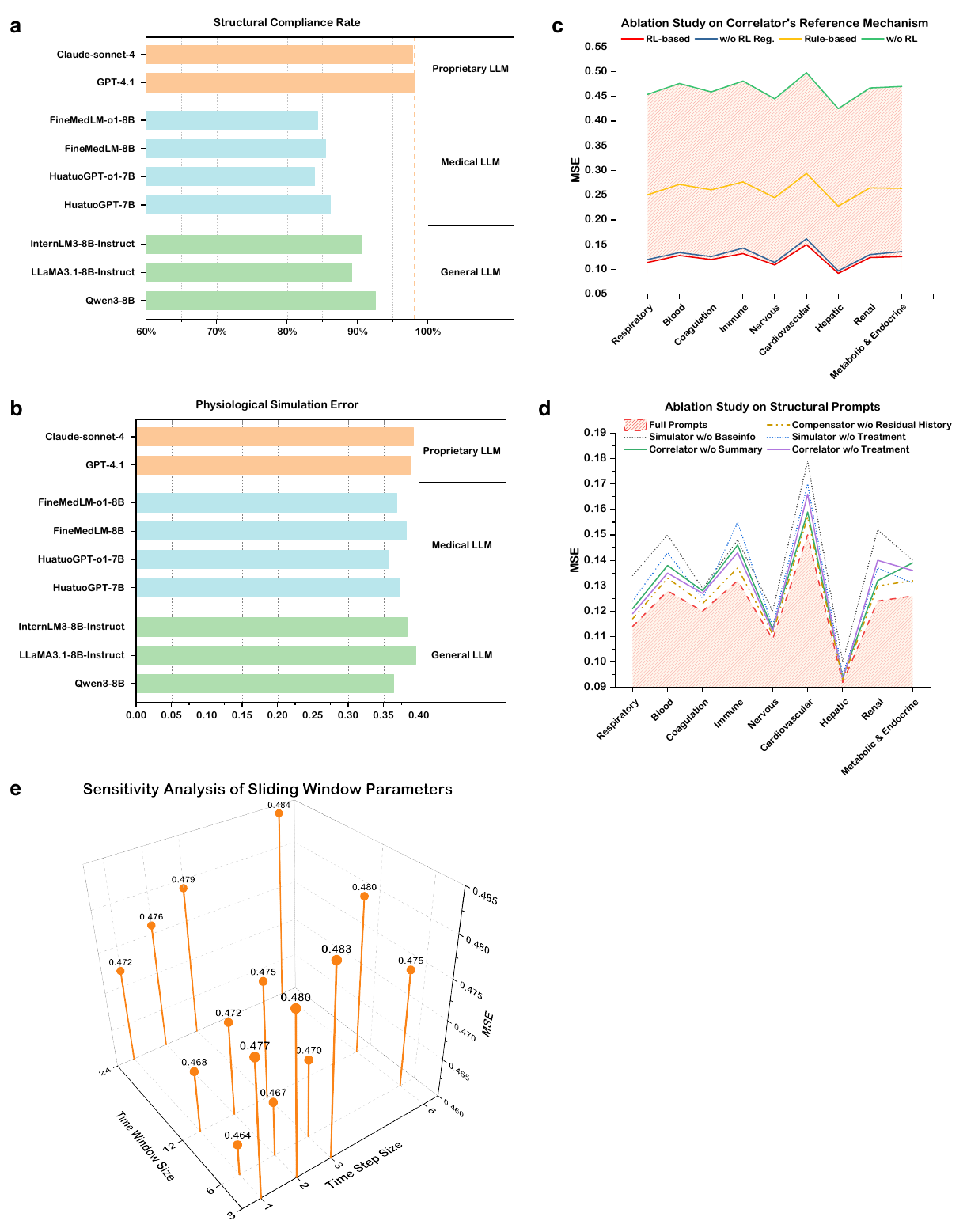}
    \caption{Ablation, sensitivity, and model evaluation experiments in Organ-Agents. 
    (\textbf{a}) Evaluation of candidate pretrained LLMs by structural compliance rate (SCR), covering general-purpose open-source models, domain-specific medical LLMs, and proprietary models. 
    (\textbf{b}) Evaluation of the same LLM candidates by physiological simulation error (PSE). 
    (\textbf{c}) Ablation on the Correlator’s reference mechanism, comparing RL-based, RL-based without regularization, rule-based, and SFT-only strategies. 
    (\textbf{d}) Structural prompt ablation across Simulator, Correlator, and Compensator modules, assessing the impact of removing specific information sources such as baseinfo, treatment, summary, or residual history. 
    (\textbf{e}) Sensitivity analysis of sliding window parameters $(w,s)$, evaluating the effect of temporal context length and stride on simulation accuracy.}
    \label{fig:app_ablation}
\end{figure}

To determine the most suitable large language model (LLM) as the backbone of our structured clinical simulator, we considered nine representative LLMs spanning three categories (Fig.~\ref{fig:app_ablation}):  
\textbf{(1) Proprietary LLMs:} GPT-4.1\cite{openai_gpt4_1} and Claude-sonnet-4\cite{anthropic_claude_sonnet_4};  
\textbf{(2) Domain-specific Medical LLMs:} FineMedLM-o1-8B\cite{yu2025finemedlm}, FineMedLM-8B\cite{yu2025finemedlm}, HuatuoGPT-o1-7B\cite{chen2024huatuogpt}, and HuatuoGPT-7B\cite{chen2024huatuogpt};  
\textbf{(3) General-purpose Open-source LLMs:} InternLM3-8B-Instruct\cite{cai2024internlm2}, LLaMA3.1-8B-Instruct\cite{grattafiori2024llama}, and Qwen3-8B\cite{qwen3}.  
These models cover a diverse spectrum of parameter scales, training objectives, and alignment strategies, enabling a comprehensive assessment of trade-offs between instruction-following ability, structural compliance, and physiological simulation accuracy.

The task involves simulating multi-indicator physiological states at a future timestamp based on structured patient-system-time prompts. Given the requirements of structural consistency and numerical precision in trajectory generation, we adopted two key evaluation metrics:

\begin{itemize}
    \item \textbf{Structural Compliance Rate (SCR)}: Measures the proportion of generated outputs that adhere strictly to the prescribed schema, including the presence and format of required \texttt{<simulation>} tags and \texttt{(value, confidence)} tuples for each system indicator.
    \item \textbf{Physiological Simulation Error (PSE)}: Quantifies the numerical accuracy of simulated values against ground truth using Mean Squared Error (MSE), averaged across all indicators and systems.
\end{itemize}

All open-source models were fine-tuned using LoRA on a subset of 600 structured training samples, while proprietary models (GPT-4.1, Claude-sonnet-4) were evaluated in zero-shot inference mode. Each model was tested on 200 held-out samples covering diverse systems and physiological states.

Figure~\ref{fig:app_ablation} (a) and (b) summarize the results for SCR and PSE, respectively. We observe that proprietary LLMs (GPT-4.1 and Claude-sonnet-4) achieve the highest SCR scores (both exceeding 97\%), confirming their strong general instruction-following capabilities. However, their PSE values remain above 0.38, indicating reduced reliability in producing stable numerical outputs. Further qualitative inspection suggests that these models tend to introduce diverse value changes even when input indicators are clinically stable, leading to unnecessary variability and ultimately degraded simulation accuracy.

Medical LLMs demonstrate a favorable trade-off between format adherence and physiological fidelity. While their SCR values are moderately lower (typically ranging from 84\% to 86\%), their PSE values are generally lower than those of general-purpose models. Notably, HuatuoGPT-o1-7B achieves the best PSE among all open-source candidates (0.358), suggesting improved alignment with physiological patterns. However, reasoning-enhanced variants (such as HuatuoGPT-o1-7B and FineMedLM-o1-8B) do not consistently outperform their standard counterparts. Their internal deliberation mechanisms, designed for diagnostic reasoning, may introduce instability during deterministic single-step generation. This phenomenon is particularly evident under constrained model scales (7–8B), where additional reasoning steps increase output variance and decoding latency without substantial benefit in accuracy.

Among general-purpose LLMs, Qwen3-8B demonstrates the most balanced performance, achieving an SCR of 92.5\% and a PSE of 0.364. This level of simulation accuracy is comparable to that of domain-specific medical LLMs, while its structural compliance is noticeably higher. In contrast, LLaMA3.1-8B-Instruct and InternLM3-8B-Instruct exhibit lower SCR values (0.892 and 0.906 respectively), and their PSE scores exceed 0.380, indicating reduced suitability for structured medical simulation tasks despite LoRA adaptation.

These findings collectively suggest that instruction-following ability in structured output formats is tightly coupled with the underlying backbone model. Most medical LLMs were optimized primarily for dialogue-based consultation scenarios, where natural language fluency is prioritized over strict schema generation. As a result, their formatting behavior heavily relies on the instruction-alignment capacity of the base model. On the other hand, simulation accuracy is more dependent on the model’s understanding of patient context and physiological patterns, which benefits from medical pretraining.

In our multi-agent simulation framework, robust instruction compliance is critical to ensuring consistency and fidelity in inter-agent communication and coordination. At the same time, simulation accuracy must be maintained to preserve the clinical validity of generated trajectories. Moreover, for practical deployment and custom adaptation across institutions, the model must be efficiently tunable and compatible with lightweight development workflows.

Taking these three factors into account, structural controllability, medical interpretability, and engineering feasibility, we select \textbf{Qwen3-8B} as the default backbone for our simulator. It provides a well-balanced trade-off between instruction alignment and physiological fidelity, while retaining the flexibility of open-source deployment and efficient LoRA-based adaptation.

\section{Ablation Study} 

\subsection{Ablation Study on Correlator's Reference Mechanism}

To comprehensively evaluate the impact of the Correlator’s reference selection, we designed an ablation experiment comparing four variants: (1) a fully adaptive RL-based Correlator trained via PPO to dynamically select relevant external systems at each step (RL-based); (2) an RL-based Correlator with RL regularization terms (entropy and sparsity penalties) removed, evaluating the effect of reduced exploration and reference parsimony (w/o RL Reg.); (3) a rule-based mechanism in which each agent receives a fixed set of references according to predefined clinical associations, independent of patient context or time (Rule-based); and (4) a single-system SFT-only baseline where each agent operates in isolation without access to any external references (w/o RL). All variants were evaluated on the same cohort with Organ-Agents using system-wise mean squared error (MSE) as the metric.

As shown in Fig.~\ref{fig:app_ablation} (c), the RL-based Correlator consistently achieves the lowest MSE across all physiological systems, demonstrating its ability to identify and integrate informative cross-system references. This advantage is observed uniformly across diverse physiological domains.

Removing RL regularization results in a mild but consistent increase in MSE, accompanied by greater fluctuations in the PPO loss curve and slower convergence. This indicates that regularization not only mitigates overfitting to transient or noisy reference patterns, but also stabilizes policy updates by constraining exploration to physiologically plausible actions.

The rule-based variant also reduces error compared to the SFT-only baseline, confirming that cross-system information benefits simulation accuracy. However, its improvements are limited by predefined selection rules that cannot adapt to atypical patient states or evolving temporal dynamics. In contrast, the RL-based Correlator achieves further gains by adaptively tailoring references to the current physiological context, rather than relying on fixed templates.

Overall, these results show that while simply adding cross-system information can be helpful, robust and generalizable multi-system simulation requires adaptive reference selection reinforced by appropriate regularization, enabling policies to refine decisions continuously based on evolving patient conditions.

\subsection{Ablation Study on Structural Prompt Modules}

To systematically evaluate the contribution of structural prompt components in Organ-Agents, we conducted a series of ablation experiments targeting each major information source or agent-specific input. Seven experimental groups were designed: (1) the full model with all prompt modules enabled (Full Prompts); (2) Simulator without baseinfo (removal of static patient characteristics, denoted as Simulator w/o baaeinfo); (3) Simulator without treatment information (removal of medication/intervention records, denoted as Simulator w/o Treatment); (4) Correlator without summary input (removal of symbolic trend summaries for cross-system referencing, denoted as Correlator w/o summary); (5) Correlator without treatment input (removal of intervention context in reference selection, denoted as Correlator w/o Treatment); and (6) Compensator without residual history (removal of prior residuals for error compensation, denoted as Compensator w/o Residual History). Each configuration was evaluated on system-wise mean squared error (MSE) across nine physiological systems, as shown in Fig.~\ref{fig:app_ablation} (d).

Removing the baseinfo module from the Simulator broadly increased simulation errors, especially in systems that are intrinsically reliant on individualized patient data due to strong baseline or comorbidity effects. For example, the cardiovascular and renal systems exhibited the largest MSE increases (from 0.150 to 0.179 and from 0.124 to 0.152, respectively), as their physiological states are tightly linked to patient-specific factors such as age, chronic disease history, and baseline organ function. In contrast, hepatic and nervous systems demonstrated relatively minor sensitivity to the loss of static baseline features (e.g., MSE increases from 0.092 to 0.100 for hepatic and from 0.109 to 0.120 for nervous), likely because their short-term dynamics are more strongly driven by acute pathophysiological changes or are less variable across patient populations in the studied cohort.

Excluding treatment information from the Simulator led to increased simulation errors, predominantly in systems that are highly responsive to clinical interventions. Cardiovascular and immune systems showed the most substantial MSE increases (from 0.150 to 0.170 and from 0.132 to 0.155, respectively), reflecting their acute modulation by medications and therapeutic actions. Other systems, such as hepatic and nervous, were less affected (e.g., hepatic: 0.092 to 0.095), likely due to a lower immediate dependence on intervention history for short-term trajectory forecasting.

Ablating symbolic summary input from the Correlator resulted in moderate but system-dependent increases in MSE, with the immune, blood, and metabolic systems most affected (e.g., immune: 0.132 to 0.146; blood: 0.128 to 0.138; metabolic: 0.126 to 0.139). This pattern highlights the importance of temporal abstraction and trend summarization in supporting accurate cross-system referencing, particularly for systems with complex, event-driven dynamics or high degrees of inter-organ modulation. Systems with more stable or less interactive trajectories, such as hepatic and nervous, were less impacted.

Similarly, removing treatment information from the Correlator primarily impacted systems that depend on context-aware referencing for acute clinical response, including cardiovascular (0.150 to 0.166), renal (0.124 to 0.140), and immune (0.132 to 0.143) systems. This result underscores the necessity of incorporating up-to-date intervention context for effective and adaptive cross-system information flow.

Finally, omitting residual history from the Compensator resulted in a modest but targeted rise in simulation errors, most notably in cardiovascular (0.150 to 0.157) and respiratory (0.114 to 0.117) systems. These findings indicate that dynamic uncertainty compensation is especially valuable in systems characterized by high volatility or cumulative error propagation over time.

Collectively, these results demonstrate that each structural prompt component imparts system-specific benefits to simulation fidelity. Tailored information, such as static baselines, intervention history, and temporal trend abstraction, is essential for accurate and generalizable multi-organ simulation.

\subsection{Sensitivity Analysis of Sliding Window Parameters}

To investigate the impact of temporal context configuration on simulation accuracy, we conducted a grid search over time window sizes $w \in \{3, 6, 12, 24\}$ hours and time step sizes $s \in \{1, 2, 3, 6\}$ hours. All models were trained using the Stage I SFT setting without reinforcement learning, ensuring that differences in mean MSE arise solely from changes in the sliding window parameters.

As shown in Fig.~\ref{fig:app_ablation} (e), the best performance was achieved with $w=6$ and $s=1$ (MSE = 0.464), indicating that a moderate temporal context coupled with fine-grained progression captures both short-term dynamics and relevant patient history without excessive noise or redundancy. While enlarging the window size generally improved performance initially by providing more historical information, excessively large windows (e.g., $w=24$) introduced abundant states that were not directly related to imminent physiological transitions. This diluted the model’s focus on critical change points and increased susceptibility to background variability, thereby degrading accuracy. Similarly, increasing the step size consistently harmed performance, as coarser strides discard intermediate observations and force the model to extrapolate over longer gaps, amplifying simulation errors. Overall, these results highlight that simulation fidelity benefits from a balanced design that preserves sufficient temporal resolution and selectively focuses on clinically relevant history, with $(w,s)=(6,1)$ offering the optimal trade-off in our ICU cohort.

\section{Extended Evaluation}
To provide a complete view of simulation performance beyond the primary results presented in the main text, we report extended quantitative and qualitative assessments of Organ-Agents across all physiological systems and all simulatable variables.

\subsection{System-Wise Simulation Accuracy}
\begin{figure}[htbp]
  \centering
  \includegraphics[width=\linewidth]{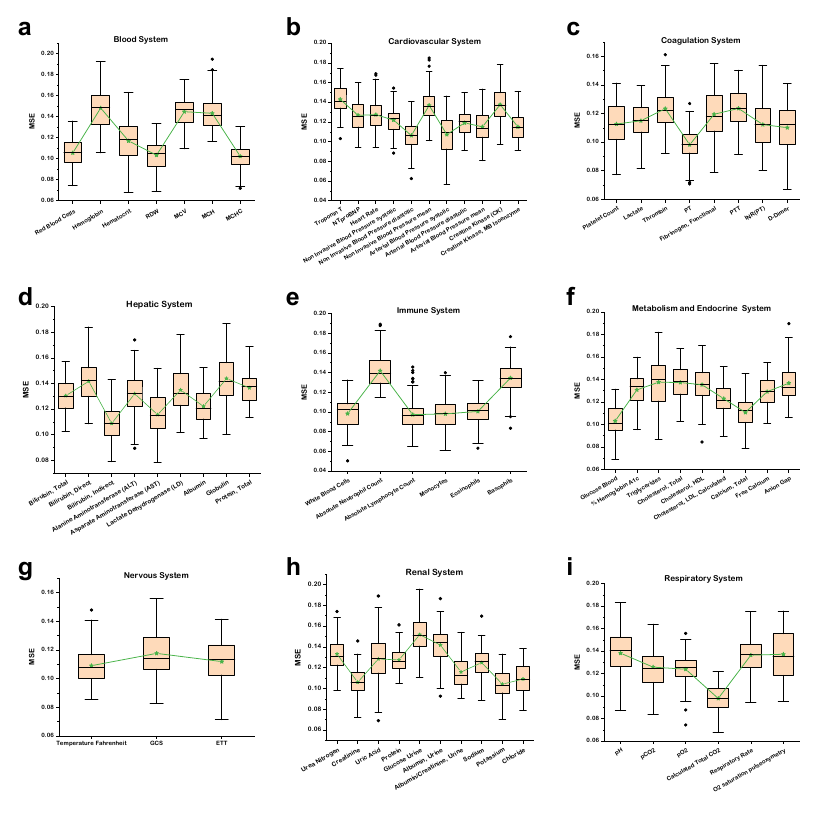}
  \caption{\textbf{System-wise simulation accuracy} 
  Mean Squared Error (MSE) of all simulatable indicators, grouped by physiological system. Each bar denotes the average MSE between predicted and true trajectories on the test set. Systems with predominantly stable dynamics, such as Renal and Hepatic, exhibit consistently low errors, whereas sporadic elevations are observed in sparsely measured or high-variance indicators (e.g., D-Dimer, NTproBNP). These results reflect system-specific variability in simulation difficulty and underscore the general accuracy of Organ-Agents across diverse physiological domains.}
  \label{fig:ap_mse}
\end{figure}

Fig.~\ref{fig:ap_mse} presents the complete distribution of simulation errors across all modeled indicators, stratified by physiological system. Each bar reflects the Mean Squared Error (MSE) between predicted and ground-truth trajectories over the test cohort. While the main text highlighted representative results, this extended analysis reveals a consistent pattern across systems: the majority of variables exhibit low simulation errors, typically below 0.15. Systems such as Renal and Cardiovascular demonstrate particularly stable performance. In contrast, slightly higher errors are observed for markers with sparse or bursty dynamics (e.g., D-Dimer, NTproBNP). These results underscore the generalizability of the model across heterogeneous physiological domains.

\subsection{Temporal Stability of Simulation Accuracy}
\begin{figure}[htbp]
  \centering
  \includegraphics[width=\linewidth]{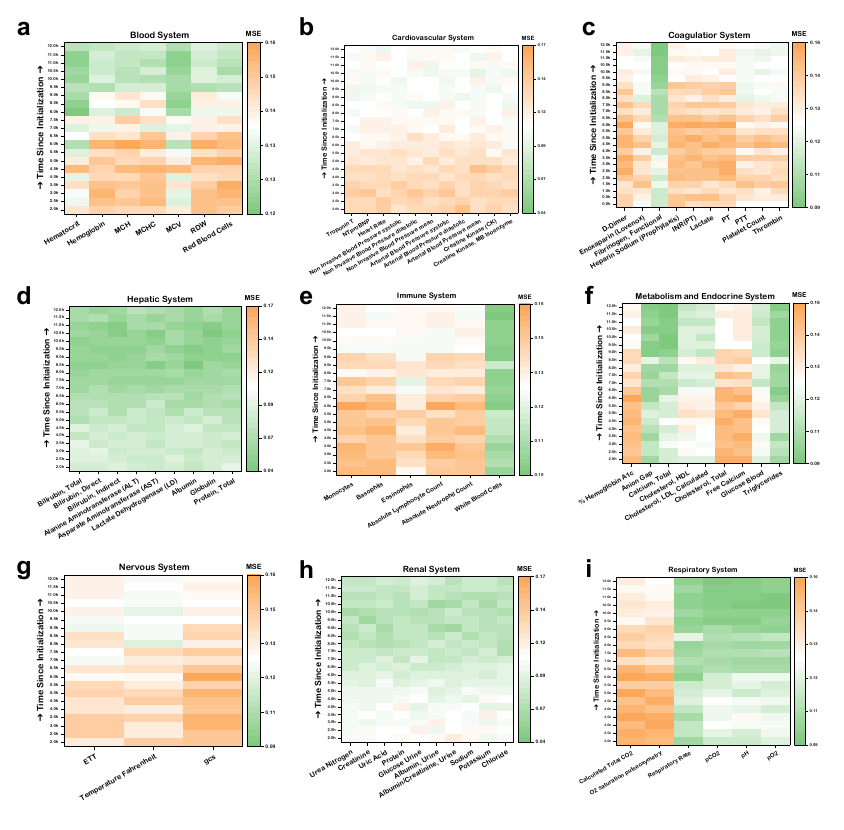}
  \caption{\textbf{Temporal stability of simulation accuracy.}
  Time-resolved MSE distributions for each physiological indicator over a 12-hour simulation window, organized by system. Error values are computed at 30-minute intervals following simulation initialization. Most systems maintain temporally stable performance, with no consistent upward drift in error magnitude. These findings indicate the robustness of Organ-Agents in sustaining accurate long-range forecasting across a wide spectrum of clinical variables.}
  \label{fig:ap_timemse}
\end{figure}

To examine the consistency of simulation quality over time, we report time-resolved MSE trends for each indicator, visualized separately by system in Fig.~\ref{fig:ap_timemse}. Errors were computed at fixed intervals up to 12 hours following initialization. Across most systems, the simulation error remains temporally stable, with only marginal fluctuations. This robustness is particularly notable in systems characterized by gradual physiological changes (e.g., Renal, Hepatic), whereas systems with faster dynamics (e.g., Respiratory, Coagulation) show modest temporal variance but no evidence of systematic drift. These findings confirm that Organ-Agents maintains stable forecasting fidelity across clinically relevant time horizons.

\subsection{Trajectory-Level Alignment with Real Patient Data}
\begin{figure*}[htbp]
  \centering
  \includegraphics[width=\textwidth]{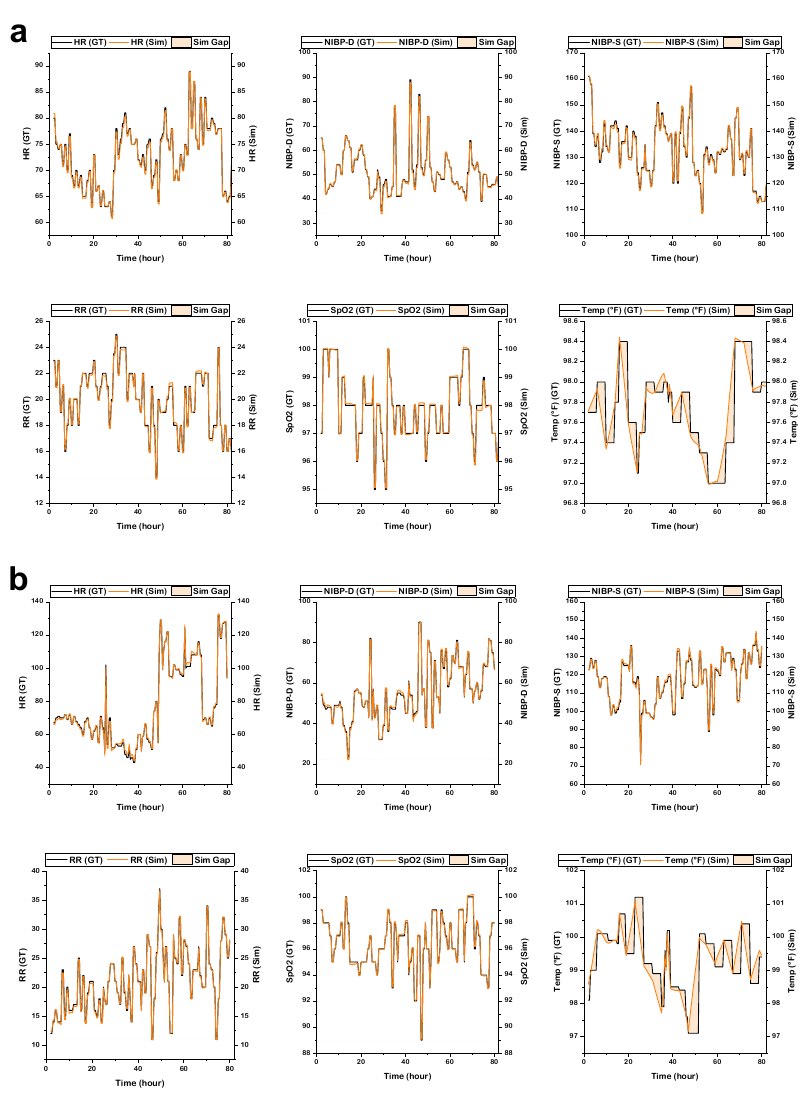}
  \caption{\textbf{Patient-specific simulation trajectories: Patients 1–2 (a–b).}
  Comparison of simulated and ground-truth temporal trajectories for two patients sampled from the test set. Each panel displays six key indicators: body temperature, oxygen saturation (SpO\textsubscript{2}), respiratory rate (RR), non-invasive systolic blood pressure (NIBP-S), diastolic blood pressure (NIBP-D), and heart rate (HR). Simulated values closely track real measurements across time, demonstrating high alignment in both trend and magnitude.}
  \label{fig:ap_simgt1}
\end{figure*}

\begin{figure*}[htbp]
  \centering
  \includegraphics[width=\textwidth]{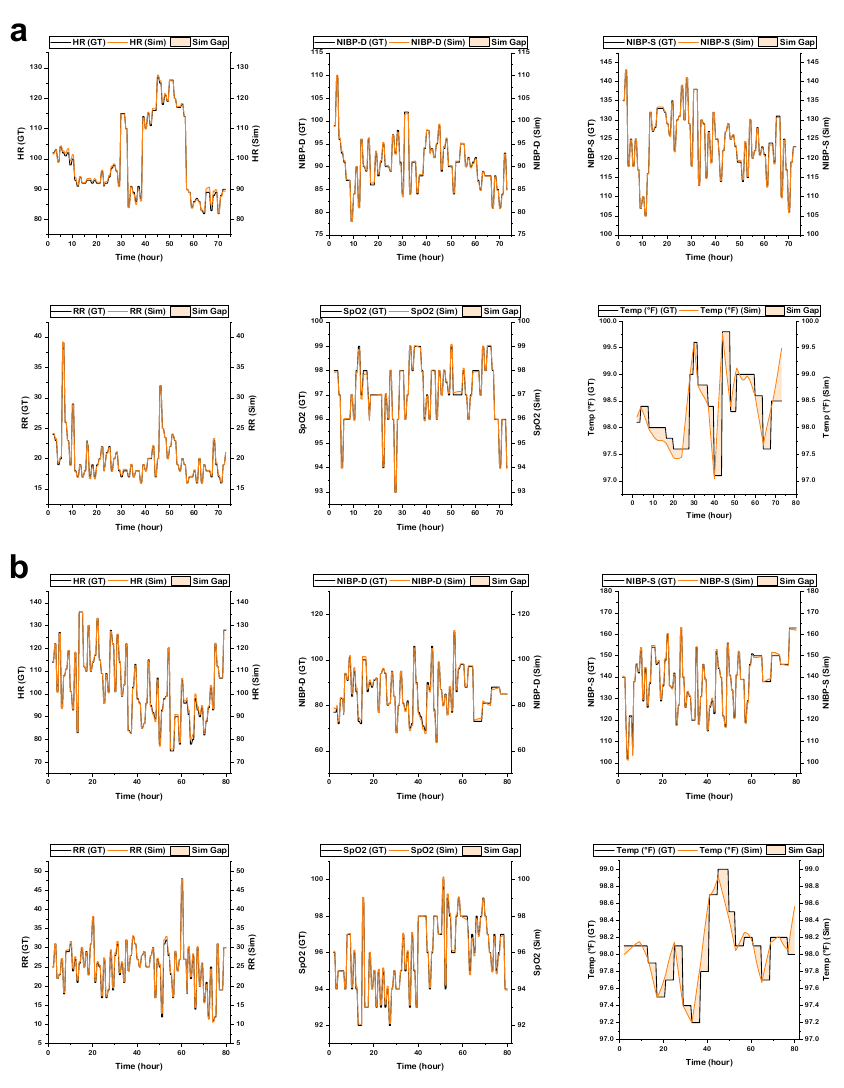}
  \caption{\textbf{Patient-specific simulation trajectories: Patients 3–4 (a–b).}
  Simulation results for two additional test-set patients, using the same six-indicator visualization as in Fig.~\ref{fig:ap_simgt1}. Despite individual variability in physiological patterns, the model accurately reproduces both gradual trends and transient shifts, indicating strong individual-level adaptability.}
  \label{fig:ap_simgt2}
\end{figure*}

\begin{figure*}[htbp]
  \centering
  \includegraphics[width=\textwidth]{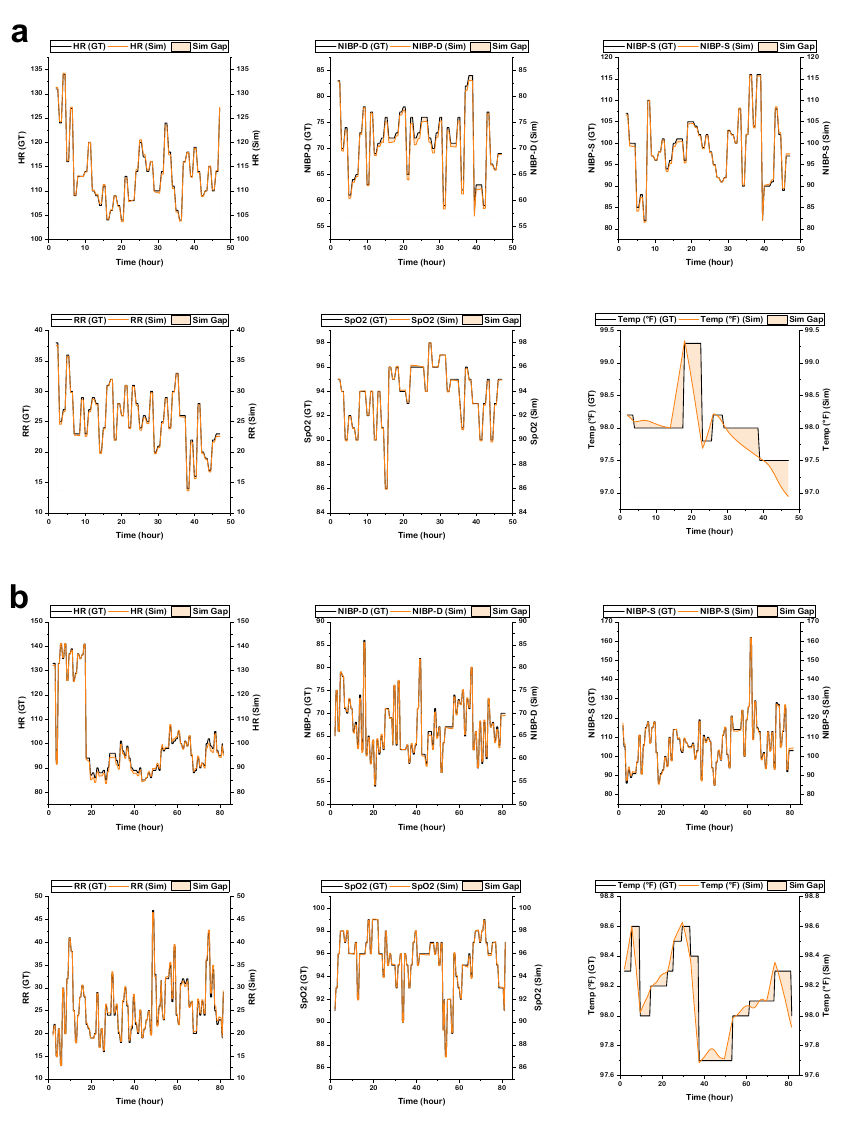}
  \caption{\textbf{Patient-specific simulation trajectories: Patients 5–6 (a–b).}
  Further examples of simulated vs. true patient trajectories, highlighting the model’s capacity to generalize across diverse clinical presentations. Simulated values exhibit tight correspondence with real-world observations across all six indicators, underscoring the consistency and realism of multi-system temporal modeling by Organ-Agents.}
  \label{fig:ap_simgt3}
\end{figure*}

To further assess the fidelity of generated trajectories, we visualize matched simulated and empirical indicator sequences for six randomly selected patients (Figs.~\ref{fig:ap_simgt1}, \ref{fig:ap_simgt2}, \ref{fig:ap_simgt3}). For each patient, six commonly monitored indicators, temperature (Temp $^\circ F$), oxygen saturation (SpO\textsubscript{2}), respiratory rate (RR), non-invasive systolic and diastolic blood pressure (NIBP-S and NIBP-D), and heart rate (HR), are plotted over time. In all cases, the simulated curves closely track the temporal trends and inflection points observed in ground-truth data. Both slow-changing and high-variability signals are well captured, and simulation gaps remain consistently narrow. These qualitative results reinforce the quantitative evidence and demonstrate the model’s capacity to reproduce individual-level physiological trajectories with high clinical plausibility.

\section{Extended Illustration of Interaction Structures} 

\subsection{Structured Prompting Framework for Organ-Agents} \label{promts}
This section presents the structured prompts that govern each specialized agent within Organ-Agents. These templates formalize system-specific reasoning tasks, enable modular learning signals, and support traceable simulation behavior across physiological domains.

\paragraph{Simulator (Stage I ): Base Simulation via System-Specific Prompts.} 
As shown in Fig.~\ref{sft_template}, the Simulator performs forward simulation based solely on base information, local temporal sequences, and recent interventions. It generates variable-wise simulations for the selected physiological system. 

\paragraph{Analyzer: Symbolic Trend Summary of Observed Sequences.} 
As shown in Fig.~\ref{analyzer_template}, the Analyzer summarizes recent temporal trends in symbolic form, supporting human-readable interpretation of system dynamics.

\paragraph{Correlator: Identification of Relevant Cross-System Indicators.} 
To simulate inter-system interaction, this agent selects relevant variables outside the target system to inform downstream generation, as shown in Fig.~\ref{corr_template}.

\paragraph{Simulator (Stage II): Reinforced Simulation with Multisystem Context.} 
As shown in Fig.~\ref{ppo_template}, the Simulator (Stage II) integrates reference data into the simulation process to enhance realism and personalization.

\paragraph{Compensator: Post-Hoc Uncertainty Adjustment.} 
When low-confidence simulations are identified, the Compensator attempts correction by analyzing short-term residual patterns, as shown in Fig.~\ref{resi_template}. 

\begin{figure}[h!]
\begin{lstlisting}
Input:
    <baseinfo>
        Patient ID 36591946 is a 53.0-year-old female, weighing 111.0kg and standing at 178.0cm tall, with a BMI of 
        35.03, indicating obesity and a body surface area (BSA) of 2.42 m2. The patient has a history of Obesity, 
        Diabetes, Hyperlipidemia, Heart failure, Ckd. The patient has no smoking and no drinking habit, and she has 
        Medicare insurance coverage. She resides in the Europe region and is single. Her ICU type is Cardiac ICU.
    <system=Respiratory>
    <ICU Time=7.5~10.0h>
        Respiratory.pH: [7.29, 7.29, 7.29, 7.32, 7.32, 7.32]
        Respiratory.pCO2: [46.0, 46.0, 46.0, 33.0, 33.0, 31.0]
        Respiratory.pO2: [94.0, 94.0, 94.0, 158.0, 158.0, 128.0]
        Respiratory.Calculated Total CO2: [23.0, 23.0, 23.0, 18.0, 18.0, 17.0]
        Respiratory.Respiratory Rate: [18.0, 18.0, 22.5, 18.0, 18.0, 18.5]
        Respiratory.O2 saturation pulseoxymetry: [99.5, 100.0, 100.0, 100.0, 99.0, 100.0]
    <treatment>
        medcine.Propofol: [9, 35.000]
        medcine.Epinephrine: [9, 0.028], [14, 0.038]
        medcine.Phenylephrine: [10, 1.200], [11, 1.000], [12, 1.400]
    Please predict each variable of Respiratory system in the format:
    <simulation>
        Respiratory.var: (value, confidence)
        Respiratory.var: (value, confidence)
        ...
    </simulation>

Output:
    <simulation>
        Respiratory.pH: (7.32, 0.98)
        Respiratory.pCO2: (30.0, 0.86)
        Respiratory.pO2: (120.0, 0.79)
        Respiratory.Calculated Total CO2: (16.5, 0.83)
        Respiratory.Respiratory Rate: (18.0, 0.92)
        Respiratory.O2 saturation pulseoxymetry: (99.5, 0.83)
    <\simulation>
\end{lstlisting}
\caption{Prompt and output format of Simulator (Stage I). This forms the base of variable-wise forward simulation.}
\label{sft_template}
\end{figure}

\begin{figure}[h!]
\begin{lstlisting}
Input:
    <system=Respiratory>
    <ICU Time=7.5~10.0h>
        Respiratory.pH: [7.29, 7.29, 7.29, 7.32, 7.32, 7.32]
        Respiratory.pCO2: [46.0, 46.0, 46.0, 33.0, 33.0, 31.0]
        Respiratory.pO2: [94.0, 94.0, 94.0, 158.0, 158.0, 128.0]
        Respiratory.Calculated Total CO2: [23.0, 23.0, 23.0, 18.0, 18.0, 17.0]
        Respiratory.Respiratory Rate: [18.0, 18.0, 22.5, 18.0, 18.0, 18.5]
        Respiratory.O2 saturation pulseoxymetry: [99.5, 100.0, 100.0, 100.0, 99.0, 100.0]
    <sum-his>
        T=4.5, Respiratory.Respiratory Rate remain stable at 18.0; Respiratory.O2 saturation pulseoxymetry remain stable at 86.5;
        T=5.0, Respiratory.Respiratory Rate fall at 18.0;
        T=5.5, Respiratory.O2 saturation pulseoxymetry rise at 98.0;
        T=6.0, Respiratory.Respiratory Rate remain stable at 18.0;
        T=7.5, Respiratory.pH fluctuate at 7.29;
        T=8.5, Respiratory.pH remain stable at 7.29; Respiratory.O2 saturation pulseoxymetry fluctuate at 100.0; Respiratory.Respiratory Rate rise at 22.5; 
        T=9.0, Respiratory.pH rise at 7.32; Respiratory.pCO2 fall at 33.0; Respiratory.pO2 rise at 158.0; Respiratory.Calculated Total CO2 fall at 18.0; Respiratory.Respiratory Rate fall at 18.0; Respiratory.O2 saturation pulseoxymetry remain stable at 100.0;
        T=9.5, Respiratory.O2 saturation pulseoxymetry fall at 99.0;
    </sum-his>
    Please summarize the trend for each variable up to the current time, 10 h. For each variable, choose one of the following event types: [rise, fall, fluctuate, no change]. The summary should be in the format:
   <summary>
       T=10, Respiratory.variable: [event type] to value; Respiratory.variable: [event type] to value; ...;
   </summary>

Output:
    <summary>
        T=10, Respiratory.pH remain stable at 7.32; Respiratory.pCO2 fall at 31.0; Respiratory.pO2 fall at 128.0; Respiratory.Calculated Total CO2 fall at 17.0; Respiratory.Respiratory Rate rise at 18.5; Respiratory.O2 saturation pulseoxymetry rise at 100.0;
    </summary>
\end{lstlisting}
\caption{Prompt and output format for the Analyzer module. It produces symbolic summaries of variable-level temporal patterns.}
\label{analyzer_template}
\end{figure}

\begin{figure}[h!]
\begin{lstlisting}
Input:
    <system=Respiratory>
    <ICU Time=7.5~10.0h>
        Respiratory.pH: [7.29, 7.29, 7.29, 7.32, 7.32, 7.32]
        Respiratory.pCO2: [46.0, 46.0, 46.0, 33.0, 33.0, 31.0]
        Respiratory.pO2: [94.0, 94.0, 94.0, 158.0, 158.0, 128.0]
        Respiratory.Calculated Total CO2: [23.0, 23.0, 23.0, 18.0, 18.0, 17.0]
        Respiratory.Respiratory Rate: [18.0, 18.0, 22.5, 18.0, 18.0, 18.5]
        Respiratory.O2 saturation pulseoxymetry: [99.5, 100.0, 100.0, 100.0, 99.0, 100.0]
    <summary>
        T=4.5, Respiratory.Respiratory Rate remain stable at 18.0; Respiratory.O2 saturation pulseoxymetry remain stable at 86.5;
        T=5.0, Respiratory.Respiratory Rate fall at 18.0;
        T=5.5, Respiratory.O2 saturation pulseoxymetry rise at 98.0;
        T=6.0, Respiratory.Respiratory Rate remain stable at 18.0;
        T=7.5, Respiratory.pH fluctuate at 7.29;
        T=8.5, Respiratory.pH remain stable at 7.29; Respiratory.O2 saturation pulseoxymetry fluctuate at 100.0; Respiratory.Respiratory Rate rise at 22.5; 
        T=9.0, Respiratory.pH rise at 7.32; Respiratory.pCO2 fall at 33.0; Respiratory.pO2 rise at 158.0; Respiratory.Calculated Total CO2 fall at 18.0; Respiratory.Respiratory Rate fall at 18.0; Respiratory.O2 saturation pulseoxymetry remain stable at 100.0;
        T=9.5, Respiratory.O2 saturation pulseoxymetry fall at 99.0;
        T=10, Respiratory.pH remain stable at 7.32; Respiratory.pCO2 fall at 31.0; Respiratory.pO2 fall at 128.0; Respiratory.Calculated Total CO2 fall at 17.0; Respiratory.Respiratory Rate rise at 18.5; Respiratory.O2 saturation pulseoxymetry rise at 100.0;
    <treatment>
        medcine.Propofol: [9, 35.000]
        medcine.Epinephrine: [9, 0.028], [14, 0.038]
        medcine.Phenylephrine: [10, 1.200], [11, 1.000], [12, 1.400]
    <candidate>
        (Potentially referenced variables from external systems.)
    </candidate>
    Please select the most relevant variables from the candidate list as references for analyzing the current Respiratory 
    system. List the selected variables in the reference block, one per line, excluding those from the Respiratory system. 
    <reference>
        system1.var1
        system2.var2
        ...
    </reference>

Output: (Appended with current states of refereced indicators)
    <reference>
        Coagulation.Platelet Count: [230.0, 230.0, 230.0, 230.0, 230.0, 245.0]
        Coagulation.Lactate: [1.1, 1.1, 1.1, 1.1, 1.1, 1.0]
        Coagulation.Thrombin: [14.1, 14.1, 14.1, 14.1, 14.1, 14.1]
        Coagulation.PT: [13.8, 13.8, 13.8, 13.8, 13.8, 13.8]
        Coagulation.Fibrinogen, Functional: [159.0, 159.0, 159.0, 159.0, 159.0, 159.0]
        Coagulation.PTT: [33.7, 33.7, 33.7, 33.7, 33.7, 33.7]
        Coagulation.INR(PT): [1.3, 1.3, 1.3, 1.3, 1.3, 1.3]
        Coagulation.D-Dimer: [737.0, 737.0, 737.0, 737.0, 737.0, 737.0]
        Immune.White Blood Cells: [19.2, 19.2, 19.2, 19.2, 19.2, 19.2]
        Cardiovascular.Heart Rate: [84.0, 84.0, 83.5, 84.0, 84.0, 84.0]
        Cardiovascular.Non Invasive Blood Pressure systolic: [134.0, 134.0, 134.0, 134.0, 134.0, 134.0]
        Cardiovascular.Non Invasive Blood Pressure diastolic: [77.0, 77.0, 77.0, 77.0, 77.0, 77.0]
        Cardiovascular.Non Invasive Blood Pressure mean: [90.0, 90.0, 90.0, 90.0, 90.0, 90.0]
        Cardiovascular.Arterial Blood Pressure systolic: [113.5, 112.0, 103.5, 101.0, 109.0, 112.0]
        Cardiovascular.Arterial Blood Pressure diastolic: [56.5, 55.0, 52.5, 53.0, 55.0, 56.5]
        Cardiovascular.Arterial Blood Pressure mean: [74.5, 72.0, 67.5, 67.5, 71.0, 72.5]
        Renal.Urea Nitrogen: [54.0, 54.0, 54.0, 54.0, 54.0, 54.0]
        Renal.Creatinine: [3.8, 3.8, 3.8, 3.8, 3.8, 3.8]
        Metabolism and endocrine.Glucose Blood: [130.0, 130.0, 130.0, 130.0, 130.0, 130.0]
        Metabolism and endocrine.Triglycerides: [79.0, 79.0, 79.0, 79.0, 79.0, 79.0]
        Metabolism and endocrine.Cholesterol, Total: [155.0, 155.0, 155.0, 155.0, 155.0, 155.0]
        Metabolism and endocrine.Cholesterol, HDL: [78.0, 78.0, 78.0, 78.0, 78.0, 78.0]
        Metabolism and endocrine.Cholesterol, LDL, Calculated: [61.0, 61.0, 61.0, 61.0, 61.0, 61.0]
    </reference>
\end{lstlisting}
\caption{Prompt and output format for the Correlator module. It selects relevant non-target-system variables for downstream simulation support.}
\label{corr_template}
\end{figure}

\begin{figure}[h!]
\begin{lstlisting}
Input:
    <baseinfo>
        Patient ID 36591946 is a 53.0-year-old female, weighing 111.0kg and standing at 178.0cm tall, with a BMI of 
        35.03, indicating obesity and a body surface area (BSA) of 2.42 m2. The patient has a history of Obesity, 
        Diabetes, Hyperlipidemia, Heart failure, Ckd. The patient has no smoking and no drinking habit, and she has 
        Medicare insurance coverage. She resides in the Europe region and is single. Her ICU type is Cardiac ICU.
    <system=Respiratory>
    <ICU Time=7.5~10.0h>
        Respiratory.pH: [7.29, 7.29, 7.29, 7.32, 7.32, 7.32]
        Respiratory.pCO2: [46.0, 46.0, 46.0, 33.0, 33.0, 31.0]
        Respiratory.pO2: [94.0, 94.0, 94.0, 158.0, 158.0, 128.0]
        Respiratory.Calculated Total CO2: [23.0, 23.0, 23.0, 18.0, 18.0, 17.0]
        Respiratory.Respiratory Rate: [18.0, 18.0, 22.5, 18.0, 18.0, 18.5]
        Respiratory.O2 saturation pulseoxymetry: [99.5, 100.0, 100.0, 100.0, 99.0, 100.0]
    <treatment>
        medcine.Propofol: [9, 35.000]
        medcine.Epinephrine: [9, 0.028], [14, 0.038]
        medcine.Phenylephrine: [10, 1.200], [11, 1.000], [12, 1.400]
    <reference>
        Coagulation.Platelet Count: [230.0, 230.0, 230.0, 230.0, 230.0, 245.0]
        Coagulation.Lactate: [1.1, 1.1, 1.1, 1.1, 1.1, 1.0]
        Coagulation.Thrombin: [14.1, 14.1, 14.1, 14.1, 14.1, 14.1]
        Coagulation.PT: [13.8, 13.8, 13.8, 13.8, 13.8, 13.8]
        Coagulation.Fibrinogen, Functional: [159.0, 159.0, 159.0, 159.0, 159.0, 159.0]
        Coagulation.PTT: [33.7, 33.7, 33.7, 33.7, 33.7, 33.7]
        Coagulation.INR(PT): [1.3, 1.3, 1.3, 1.3, 1.3, 1.3]
        Coagulation.D-Dimer: [737.0, 737.0, 737.0, 737.0, 737.0, 737.0]
        Immune.White Blood Cells: [19.2, 19.2, 19.2, 19.2, 19.2, 19.2]
        Cardiovascular.Heart Rate: [84.0, 84.0, 83.5, 84.0, 84.0, 84.0]
        Cardiovascular.Non Invasive Blood Pressure systolic: [134.0, 134.0, 134.0, 134.0, 134.0, 134.0]
        Cardiovascular.Non Invasive Blood Pressure diastolic: [77.0, 77.0, 77.0, 77.0, 77.0, 77.0]
        Cardiovascular.Non Invasive Blood Pressure mean: [90.0, 90.0, 90.0, 90.0, 90.0, 90.0]
        Cardiovascular.Arterial Blood Pressure systolic: [113.5, 112.0, 103.5, 101.0, 109.0, 112.0]
        Cardiovascular.Arterial Blood Pressure diastolic: [56.5, 55.0, 52.5, 53.0, 55.0, 56.5]
        Cardiovascular.Arterial Blood Pressure mean: [74.5, 72.0, 67.5, 67.5, 71.0, 72.5]
        Renal.Urea Nitrogen: [54.0, 54.0, 54.0, 54.0, 54.0, 54.0]
        Renal.Creatinine: [3.8, 3.8, 3.8, 3.8, 3.8, 3.8]
        Metabolism and endocrine.Glucose Blood: [130.0, 130.0, 130.0, 130.0, 130.0, 130.0]
        Metabolism and endocrine.Triglycerides: [79.0, 79.0, 79.0, 79.0, 79.0, 79.0]
        Metabolism and endocrine.Cholesterol, Total: [155.0, 155.0, 155.0, 155.0, 155.0, 155.0]
        Metabolism and endocrine.Cholesterol, HDL: [78.0, 78.0, 78.0, 78.0, 78.0, 78.0]
        Metabolism and endocrine.Cholesterol, LDL, Calculated: [61.0, 61.0, 61.0, 61.0, 61.0, 61.0]
    </reference>
    Please predict each variable of Respiratory system in the format:
    <simulation>
        Respiratory.var: (value, confidence)
        Respiratory.var: (value, confidence)
        ...
    </simulation>

Output:
    <simulation>
        Respiratory.pH: (7.32, 0.88)
        Respiratory.pCO2: (31.0, 0.92)
        Respiratory.pO2: (128.0, 0.99)
        Respiratory.Calculated Total CO2: (17.0, 0.97)
        Respiratory.Respiratory Rate: (18.5, 1.0)
        Respiratory.O2 saturation pulseoxymetry: (99.5, 0.95)
    <\simulation>
\end{lstlisting}
\caption{Augmented prompt and output format for Stage II simulation. It incorporates system-wise interactions to refine simulations.}
\label{ppo_template}
\end{figure}

\begin{figure}[h!]
\begin{lstlisting}
Input:
    <system=Respiratory>
    <ICU Time=7.5~10.0h>
        Respiratory.pH: [7.29, 7.29, 7.29, 7.32, 7.32, 7.32]
        Respiratory.pCO2: [46.0, 46.0, 46.0, 33.0, 33.0, 31.0]
        Respiratory.pO2: [94.0, 94.0, 94.0, 158.0, 158.0, 128.0]
        Respiratory.Calculated Total CO2: [23.0, 23.0, 23.0, 18.0, 18.0, 17.0]
        Respiratory.Respiratory Rate: [18.0, 18.0, 22.5, 18.0, 18.0, 18.5]
        Respiratory.O2 saturation pulseoxymetry: [99.5, 100.0, 100.0, 100.0, 99.0, 100.0]
    <simulation>
        Respiratory.pH: (7.32, 0.88)
        Respiratory.pCO2: (31.0, 0.92)
        Respiratory.pO2: (128.0, 0.99)
        Respiratory.Calculated Total CO2: (17.0, 0.97)
        Respiratory.Respiratory Rate: (18.5, 1.0)
        Respiratory.O2 saturation pulseoxymetry: (99.5, 0.95)
    <\simulation>
    <res-his>
        Respiratory.pH: [null, null, 0.02, null, null, null]
        Respiratory.pCO2: [null, -2.0, null, null, 0.5, null]
        Respiratory.pO2: [null, null, null, null, null, null]
        Respiratory.Calculated Total CO2: [null, null, null, null, -0.5, -0.5]
        Respiratory.Respiratory Rate: [null, null, null, null, null, null]
        Respiratory.O2 saturation pulseoxymetry: [null, null, 0.2, null, null, null]
    <res-his>
    Please provide residuals for variables with uncertain simulations (confidence < 0.8), using the following format ('null' if not applicable):
    <residual>
        Respiratory.var: (residual)
        Respiratory.var: (residual)
        ...
    </residual>

Output:
    <residual>
        Respiratory.pH: (null)
        Respiratory.pCO2: (null)
        Respiratory.pO2: (null)
        Respiratory.Calculated Total CO2: (null)
        Respiratory.Respiratory Rate: (null)
        Respiratory.O2 saturation pulseoxymetry: (null)
    <\residual>
\end{lstlisting}
\caption{Prompt and ouput format of the Residual Compensator. It detects uncertain simulations and proposes optional adjustments based on recent error traces.}
\label{resi_template}
\end{figure}

\subsection*{B.2 Interactive Interface for Executing Structured Simulation}
To complement the structured prompt logic described above, we developed a user-facing interface that enables real-time interaction with Organ-Agents. This interface encapsulates the full simulation loop: from data input to variable simulation, reference integration, and trajectory visualization, all aligned with the agentic execution workflow.

Users initiate simulation by uploading structured patient data including baseline information, temporal system indicators, treatments, and optional cross-system references. Each reasoning module described in Appendix \ref{promts} is automatically triggered in sequence. The results—such as predicted variable values with confidence scores—are displayed alongside historical trajectories, with simulated values highlighted to distinguish them from original inputs.

\begin{figure}[h!]
  \centering
  \includegraphics[width=0.95\linewidth]{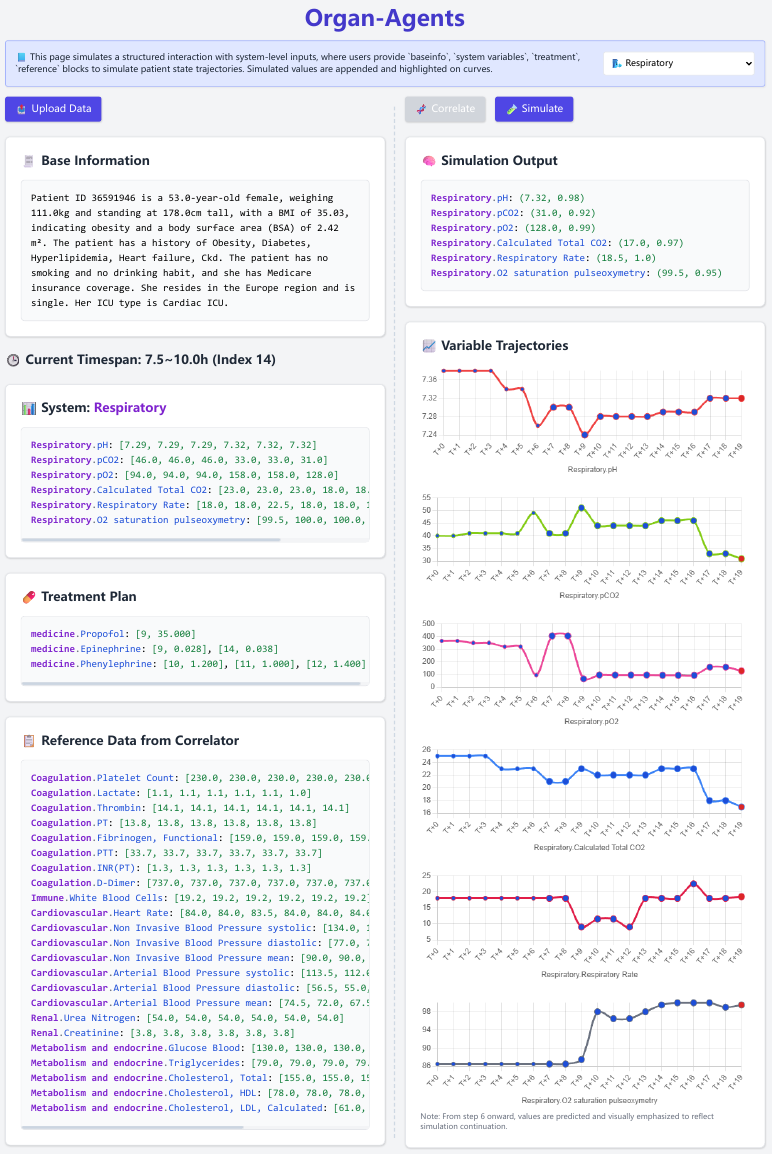}
  \caption{\textbf{Organ-Agents interface for executing system-level simulations.} The interface orchestrates all reasoning agents and visually appends simulation outputs to time-series trajectories, supporting interpretable and modular analysis of patient physiology.}
  \label{fig:simulator_ui}
\end{figure}

As shown in Figure~\ref{fig:simulator_ui}, the interface supports transparent observation of system reasoning and facilitates expert intervention if needed. Beyond passive visualization, it also enables hypothesis testing through prompt-level manipulation of inputs, allowing dynamic exploration of treatment-response pathways or physiological counterfactuals. 

\end{appendices}

\clearpage
\bibliography{sn-bibliography}% common bib file


\begin{thebibliography}{48}
\ifx \bisbn   \undefined \def \bisbn  #1{ISBN #1}\fi
\ifx \binits  \undefined \def \binits#1{#1}\fi
\ifx \bauthor  \undefined \def \bauthor#1{#1}\fi
\ifx \batitle  \undefined \def \batitle#1{#1}\fi
\ifx \bjtitle  \undefined \def \bjtitle#1{#1}\fi
\ifx \bvolume  \undefined \def \bvolume#1{\textbf{#1}}\fi
\ifx \byear  \undefined \def \byear#1{#1}\fi
\ifx \bissue  \undefined \def \bissue#1{#1}\fi
\ifx \bfpage  \undefined \def \bfpage#1{#1}\fi
\ifx \blpage  \undefined \def \blpage #1{#1}\fi
\ifx \burl  \undefined \def \burl#1{\textsf{#1}}\fi
\ifx \doiurl  \undefined \def \doiurl#1{\url{https://doi.org/#1}}\fi
\ifx \betal  \undefined \def \betal{\textit{et al.}}\fi
\ifx \binstitute  \undefined \def \binstitute#1{#1}\fi
\ifx \binstitutionaled  \undefined \def \binstitutionaled#1{#1}\fi
\ifx \bctitle  \undefined \def \bctitle#1{#1}\fi
\ifx \beditor  \undefined \def \beditor#1{#1}\fi
\ifx \bpublisher  \undefined \def \bpublisher#1{#1}\fi
\ifx \bbtitle  \undefined \def \bbtitle#1{#1}\fi
\ifx \bedition  \undefined \def \bedition#1{#1}\fi
\ifx \bseriesno  \undefined \def \bseriesno#1{#1}\fi
\ifx \blocation  \undefined \def \blocation#1{#1}\fi
\ifx \bsertitle  \undefined \def \bsertitle#1{#1}\fi
\ifx \bsnm \undefined \def \bsnm#1{#1}\fi
\ifx \bsuffix \undefined \def \bsuffix#1{#1}\fi
\ifx \bparticle \undefined \def \bparticle#1{#1}\fi
\ifx \barticle \undefined \def \barticle#1{#1}\fi
\bibcommenthead
\ifx \bconfdate \undefined \def \bconfdate #1{#1}\fi
\ifx \botherref \undefined \def \botherref #1{#1}\fi
\ifx \url \undefined \def \url#1{\textsf{#1}}\fi
\ifx \bchapter \undefined \def \bchapter#1{#1}\fi
\ifx \bbook \undefined \def \bbook#1{#1}\fi
\ifx \bcomment \undefined \def \bcomment#1{#1}\fi
\ifx \oauthor \undefined \def \oauthor#1{#1}\fi
\ifx \citeauthoryear \undefined \def \citeauthoryear#1{#1}\fi
\ifx \endbibitem  \undefined \def \endbibitem {}\fi
\ifx \bconflocation  \undefined \def \bconflocation#1{#1}\fi
\ifx \arxivurl  \undefined \def \arxivurl#1{\textsf{#1}}\fi
\csname PreBibitemsHook\endcsname

%%% 1
\bibitem[\protect\citeauthoryear{Bashan et~al.}{2012}]{bashan2012network}
\begin{barticle}
\bauthor{\bsnm{Bashan}, \binits{A.}},
\bauthor{\bsnm{Bartsch}, \binits{R.P.}},
\bauthor{\bsnm{Kantelhardt}, \binits{J.W.}},
\bauthor{\bsnm{Havlin}, \binits{S.}},
\bauthor{\bsnm{Ivanov}, \binits{P.C.}}:
\batitle{Network physiology reveals relations between network topology and physiological function}.
\bjtitle{Nature Communications}
\bvolume{3},
\bfpage{702}
(\byear{2012})
\doiurl{10.1038/ncomms1705}
\end{barticle}
\endbibitem

%%% 2
\bibitem[\protect\citeauthoryear{Lehnertz et~al.}{2020}]{lehnertz2020human}
\begin{barticle}
\bauthor{\bsnm{Lehnertz}, \binits{K.}},
\bauthor{\bsnm{Bröhl}, \binits{T.}},
\bauthor{\bsnm{Rings}, \binits{T.}}:
\batitle{The human organism as an integrated interaction network: Recent conceptual and methodological challenges}.
\bjtitle{Frontiers in Physiology}
\bvolume{11},
\bfpage{598694}
(\byear{2020})
\doiurl{10.3389/fphys.2020.598694}
\end{barticle}
\endbibitem

%%% 3
\bibitem[\protect\citeauthoryear{Hama et~al.}{2025}]{hama2025deep}
\begin{barticle}
\bauthor{\bsnm{Hama}, \binits{T.}},
\bauthor{\bsnm{Alsaleh}, \binits{M.M.}},
\bauthor{\bsnm{Allery}, \binits{F.}},
\bauthor{\bsnm{Choi}, \binits{J.W.}},
\bauthor{\bsnm{Tomlinson}, \binits{C.}},
\bauthor{\bsnm{Wu}, \binits{H.}},
\bauthor{\bsnm{Lai}, \binits{A.}},
\bauthor{\bsnm{Pontikos}, \binits{N.}},
\bauthor{\bsnm{Thygesen}, \binits{J.H.}}:
\batitle{Enhancing patient outcome prediction through deep learning with sequential diagnosis codes from structured electronic health record data: Systematic review}.
\bjtitle{Journal of Medical Internet Research}
\bvolume{27},
\bfpage{57358}
(\byear{2025})
\doiurl{10.2196/57358}
\end{barticle}
\endbibitem

%%% 4
\bibitem[\protect\citeauthoryear{Xie et~al.}{2022}]{xie2022deep}
\begin{barticle}
\bauthor{\bsnm{Xie}, \binits{F.}},
\bauthor{\bsnm{Yuan}, \binits{H.}},
\bauthor{\bsnm{Ning}, \binits{Y.}},
\bauthor{\bsnm{Ong}, \binits{M.H.E.}},
\bauthor{\bsnm{Feng}, \binits{M.}},
\bauthor{\bsnm{Hsu}, \binits{W.}},
\bauthor{\bsnm{Chakraborty}, \binits{B.}},
\bauthor{\bsnm{Liu}, \binits{N.}}:
\batitle{Deep learning for temporal data representation in electronic health records: A systematic review of challenges and methodologies}.
\bjtitle{Journal of Biomedical Informatics}
\bvolume{126},
\bfpage{103980}
(\byear{2022})
\doiurl{10.1016/j.jbi.2021.103980} .
\bcomment{Epub 2021 Dec 30}
\end{barticle}
\endbibitem

%%% 5
\bibitem[\protect\citeauthoryear{Aitia}{2024}]{aitia2024causal}
\begin{barticle}
\bauthor{\bsnm{Aitia}}:
\batitle{Causal artificial intelligence and digital twins are transforming drug discovery and development}.
\bjtitle{Nature Biopharma Dealmakers}
(\byear{2024})
\doiurl{10.1038/d43747-024-00128-1}
\end{barticle}
\endbibitem

%%% 6
\bibitem[\protect\citeauthoryear{Martinez-Velazquez et~al.}{2019}]{martinez2019cardio}
\begin{bchapter}
\bauthor{\bsnm{Martinez-Velazquez}, \binits{R.}},
\bauthor{\bsnm{D{\'i}az}, \binits{R.G.}},
\bauthor{\bsnm{Saddik}, \binits{A.E.}}:
\bctitle{Cardio twin: A digital twin of the human heart running on the edge}.
In: \bbtitle{2019 IEEE International Symposium on Medical Measurements and Applications (MeMeA)},
pp. \bfpage{1}--\blpage{6}
(\byear{2019}).
\bcomment{IEEE}
\end{bchapter}
\endbibitem

%%% 7
\bibitem[\protect\citeauthoryear{Luo et~al.}{2022}]{luo2022biogpt}
\begin{barticle}
\bauthor{\bsnm{Luo}, \binits{R.}},
\bauthor{\bsnm{Sun}, \binits{L.}},
\bauthor{\bsnm{Xia}, \binits{Y.}},
\bauthor{\bsnm{Qin}, \binits{T.}},
\bauthor{\bsnm{Zhang}, \binits{S.}},
\bauthor{\bsnm{Poon}, \binits{H.}},
\bauthor{\bsnm{Liu}, \binits{T.-Y.}}:
\batitle{Biogpt: generative pre-trained transformer for biomedical text generation and mining}.
\bjtitle{Briefings in bioinformatics}
\bvolume{23}(\bissue{6}),
\bfpage{409}
(\byear{2022})
\end{barticle}
\endbibitem

%%% 8
\bibitem[\protect\citeauthoryear{Huang et~al.}{2020}]{huang2020clinicalbertmodelingclinicalnotes}
\begin{botherref}
\oauthor{\bsnm{Huang}, \binits{K.}},
\oauthor{\bsnm{Altosaar}, \binits{J.}},
\oauthor{\bsnm{Ranganath}, \binits{R.}}:
ClinicalBERT: Modeling Clinical Notes and Predicting Hospital Readmission
(2020).
\url{https://arxiv.org/abs/1904.05342}
\end{botherref}
\endbibitem

%%% 9
\bibitem[\protect\citeauthoryear{Jiang et~al.}{2023}]{jiang2023nyutron}
\begin{barticle}
\bauthor{\bsnm{Jiang}, \binits{L.Y.}},
\bauthor{\bsnm{Liu}, \binits{X.C.}},
\bauthor{\bsnm{Nejatian}, \binits{N.P.}},
\bauthor{\bsnm{Nasir-Moin}, \binits{M.}},
\bauthor{\bsnm{Wang}, \binits{D.}},
\bauthor{\bsnm{Abidin}, \binits{A.}},
\bauthor{\bsnm{Eaton}, \binits{K.}},
\bauthor{\bsnm{al.}}:
\batitle{Health system-scale language models are all-purpose prediction engines}.
\bjtitle{Nature}
\bvolume{619}(\bissue{7969}),
\bfpage{357}--\blpage{362}
(\byear{2023})
\doiurl{10.1038/s41586-023-06160-y}
\end{barticle}
\endbibitem

%%% 10
\bibitem[\protect\citeauthoryear{Johnson et~al.}{2023}]{johnson2023mimiciv}
\begin{botherref}
\oauthor{\bsnm{Johnson}, \binits{A.E.W.}},
\oauthor{\bsnm{Bulgarelli}, \binits{L.}},
\oauthor{\bsnm{Shen}, \binits{L.}}, et al.:
{MIMIC-IV, a freely accessible electronic health record dataset}.
Scientific Data
\textbf{10}(1)
(2023)
\doiurl{10.1038/s41597-022-01899-x}
\end{botherref}
\endbibitem

%%% 11
\bibitem[\protect\citeauthoryear{Knaus et~al.}{1985}]{knaus1985apache}
\begin{barticle}
\bauthor{\bsnm{Knaus}, \binits{W.A.}},
\bauthor{\bsnm{Draper}, \binits{E.A.}},
\bauthor{\bsnm{Wagner}, \binits{D.P.}},
\bauthor{\bsnm{Zimmerman}, \binits{J.E.}}:
\batitle{Apache ii: a severity of disease classification system}.
\bjtitle{Critical care medicine}
\bvolume{13}(\bissue{10}),
\bfpage{818}--\blpage{829}
(\byear{1985})
\end{barticle}
\endbibitem

%%% 12
\bibitem[\protect\citeauthoryear{Cho et~al.}{2014}]{cho2014gru}
\begin{botherref}
\oauthor{\bsnm{Cho}, \binits{K.}},
\oauthor{\bsnm{Merrienboer}, \binits{B.}},
\oauthor{\bsnm{Gulcehre}, \binits{C.}},
\oauthor{\bsnm{Bahdanau}, \binits{D.}},
\oauthor{\bsnm{Bougares}, \binits{F.}},
\oauthor{\bsnm{Schwenk}, \binits{H.}},
\oauthor{\bsnm{Bengio}, \binits{Y.}}:
Learning phrase representations using rnn encoder-decoder for statistical machine translation.
arXiv preprint arXiv:1406.1078
(2014)
\end{botherref}
\endbibitem

%%% 13
\bibitem[\protect\citeauthoryear{Hochreiter and Schmidhuber}{1997}]{hochreiter1997long}
\begin{barticle}
\bauthor{\bsnm{Hochreiter}, \binits{S.}},
\bauthor{\bsnm{Schmidhuber}, \binits{J.}}:
\batitle{Long short-term memory}.
\bjtitle{Neural computation}
\bvolume{9}(\bissue{8}),
\bfpage{1735}--\blpage{1780}
(\byear{1997})
\end{barticle}
\endbibitem

%%% 14
\bibitem[\protect\citeauthoryear{Bai et~al.}{2018}]{bai2018empirical}
\begin{botherref}
\oauthor{\bsnm{Bai}, \binits{S.}},
\oauthor{\bsnm{Kolter}, \binits{J.Z.}},
\oauthor{\bsnm{Koltun}, \binits{V.}}:
An empirical evaluation of generic convolutional and recurrent networks for sequence modeling.
arXiv preprint arXiv:1803.01271
(2018)
\end{botherref}
\endbibitem

%%% 15
\bibitem[\protect\citeauthoryear{Rumelhart et~al.}{1986}]{rumelhart1986mlp}
\begin{barticle}
\bauthor{\bsnm{Rumelhart}, \binits{D.E.}},
\bauthor{\bsnm{Hinton}, \binits{G.E.}},
\bauthor{\bsnm{Williams}, \binits{R.J.}}:
\batitle{Learning representations by back-propagating errors}.
\bjtitle{Nature}
\bvolume{323}(\bissue{6088}),
\bfpage{533}--\blpage{536}
(\byear{1986})
\end{barticle}
\endbibitem

%%% 16
\bibitem[\protect\citeauthoryear{Chen and Guestrin}{2016}]{chen2016xgboost}
\begin{bchapter}
\bauthor{\bsnm{Chen}, \binits{T.}},
\bauthor{\bsnm{Guestrin}, \binits{C.}}:
\bctitle{Xgboost: A scalable tree boosting system}.
In: \bbtitle{Proceedings of the 22nd ACM SIGKDD International Conference on Knowledge Discovery and Data Mining},
pp. \bfpage{785}--\blpage{794}
(\byear{2016})
\end{bchapter}
\endbibitem

%%% 17
\bibitem[\protect\citeauthoryear{Breiman}{2001}]{breiman2001random}
\begin{barticle}
\bauthor{\bsnm{Breiman}, \binits{L.}}:
\batitle{Random forests}.
\bjtitle{Machine learning}
\bvolume{45}(\bissue{1}),
\bfpage{5}--\blpage{32}
(\byear{2001})
\end{barticle}
\endbibitem

%%% 18
\bibitem[\protect\citeauthoryear{McCallum et~al.}{1998}]{mccallum1998comparison}
\begin{bchapter}
\bauthor{\bsnm{McCallum}, \binits{A.}},
\bauthor{\bsnm{Nigam}, \binits{K.}}, \betal:
\bctitle{A comparison of event models for naive bayes text classification}.
In: \bbtitle{AAAI-98 Workshop on Learning for Text Categorization},
vol. \bseriesno{752},
pp. \bfpage{41}--\blpage{48}
(\byear{1998}).
\bcomment{Madison, WI}
\end{bchapter}
\endbibitem

%%% 19
\bibitem[\protect\citeauthoryear{Singhal et~al.}{2023}]{singhal2023large}
\begin{barticle}
\bauthor{\bsnm{Singhal}, \binits{K.}},
\bauthor{\bsnm{Azizi}, \binits{S.}},
\bauthor{\bsnm{Tu}, \binits{T.}},
\bauthor{\bsnm{Mahdavi}, \binits{S.S.}},
\bauthor{\bsnm{Wei}, \binits{J.}},
\bauthor{\bsnm{Chung}, \binits{H.W.}},
\bauthor{\bsnm{Scales}, \binits{N.}},
\bauthor{\bsnm{Tanwani}, \binits{A.}},
\bauthor{\bsnm{Cole-Lewis}, \binits{H.}},
\bauthor{\bsnm{Pfohl}, \binits{S.}}, \betal:
\batitle{Large language models encode clinical knowledge}.
\bjtitle{Nature}
\bvolume{620}(\bissue{7972}),
\bfpage{172}--\blpage{180}
(\byear{2023})
\end{barticle}
\endbibitem

%%% 20
\bibitem[\protect\citeauthoryear{Goh et~al.}{2024}]{goh2024large}
\begin{barticle}
\bauthor{\bsnm{Goh}, \binits{E.}},
\bauthor{\bsnm{Gallo}, \binits{R.}},
\bauthor{\bsnm{Hom}, \binits{J.}},
\bauthor{\bsnm{Strong}, \binits{E.}},
\bauthor{\bsnm{Weng}, \binits{Y.}},
\bauthor{\bsnm{Kerman}, \binits{H.}},
\bauthor{\bsnm{Cool}, \binits{J.A.}},
\bauthor{\bsnm{Kanjee}, \binits{Z.}},
\bauthor{\bsnm{Parsons}, \binits{A.S.}},
\bauthor{\bsnm{Ahuja}, \binits{N.}}, \betal:
\batitle{Large language model influence on diagnostic reasoning: a randomized clinical trial}.
\bjtitle{JAMA Network Open}
\bvolume{7}(\bissue{10}),
\bfpage{2440969}--\blpage{2440969}
(\byear{2024})
\end{barticle}
\endbibitem

%%% 21
\bibitem[\protect\citeauthoryear{Van~Veen et~al.}{2023}]{van2023clinical}
\begin{botherref}
\oauthor{\bsnm{Van~Veen}, \binits{D.}},
\oauthor{\bsnm{Van~Uden}, \binits{C.}},
\oauthor{\bsnm{Blankemeier}, \binits{L.}},
\oauthor{\bsnm{Delbrouck}, \binits{J.-B.}},
\oauthor{\bsnm{Aali}, \binits{A.}},
\oauthor{\bsnm{Bluethgen}, \binits{C.}},
\oauthor{\bsnm{Pareek}, \binits{A.}},
\oauthor{\bsnm{Polacin}, \binits{M.}},
\oauthor{\bsnm{Reis}, \binits{E.P.}},
\oauthor{\bsnm{Seehofnerova}, \binits{A.}}, et al.:
Clinical text summarization: adapting large language models can outperform human experts.
Research square,
3
(2023)
\end{botherref}
\endbibitem

%%% 22
\bibitem[\protect\citeauthoryear{Ayers et~al.}{2023}]{ayers2023comparing}
\begin{barticle}
\bauthor{\bsnm{Ayers}, \binits{J.W.}},
\bauthor{\bsnm{Poliak}, \binits{A.}},
\bauthor{\bsnm{Dredze}, \binits{M.}},
\bauthor{\bsnm{Leas}, \binits{E.C.}},
\bauthor{\bsnm{Zhu}, \binits{Z.}},
\bauthor{\bsnm{Kelley}, \binits{J.B.}},
\bauthor{\bsnm{Faix}, \binits{D.J.}},
\bauthor{\bsnm{Goodman}, \binits{A.M.}},
\bauthor{\bsnm{Longhurst}, \binits{C.A.}},
\bauthor{\bsnm{Hogarth}, \binits{M.}}, \betal:
\batitle{Comparing physician and artificial intelligence chatbot responses to patient questions posted to a public social media forum}.
\bjtitle{JAMA internal medicine}
\bvolume{183}(\bissue{6}),
\bfpage{589}--\blpage{596}
(\byear{2023})
\end{barticle}
\endbibitem

%%% 23
\bibitem[\protect\citeauthoryear{Moor et~al.}{2023}]{moor2023med}
\begin{bchapter}
\bauthor{\bsnm{Moor}, \binits{M.}},
\bauthor{\bsnm{Huang}, \binits{Q.}},
\bauthor{\bsnm{Wu}, \binits{S.}},
\bauthor{\bsnm{Yasunaga}, \binits{M.}},
\bauthor{\bsnm{Dalmia}, \binits{Y.}},
\bauthor{\bsnm{Leskovec}, \binits{J.}},
\bauthor{\bsnm{Zakka}, \binits{C.}},
\bauthor{\bsnm{Reis}, \binits{E.P.}},
\bauthor{\bsnm{Rajpurkar}, \binits{P.}}:
\bctitle{Med-flamingo: a multimodal medical few-shot learner}.
In: \bbtitle{Machine Learning for Health (ML4H)},
pp. \bfpage{353}--\blpage{367}
(\byear{2023}).
\bcomment{PMLR}
\end{bchapter}
\endbibitem

%%% 24
\bibitem[\protect\citeauthoryear{Li et~al.}{2023}]{li2023llava}
\begin{barticle}
\bauthor{\bsnm{Li}, \binits{C.}},
\bauthor{\bsnm{Wong}, \binits{C.}},
\bauthor{\bsnm{Zhang}, \binits{S.}},
\bauthor{\bsnm{Usuyama}, \binits{N.}},
\bauthor{\bsnm{Liu}, \binits{H.}},
\bauthor{\bsnm{Yang}, \binits{J.}},
\bauthor{\bsnm{Naumann}, \binits{T.}},
\bauthor{\bsnm{Poon}, \binits{H.}},
\bauthor{\bsnm{Gao}, \binits{J.}}:
\batitle{Llava-med: Training a large language-and-vision assistant for biomedicine in one day}.
\bjtitle{Advances in Neural Information Processing Systems}
\bvolume{36},
\bfpage{28541}--\blpage{28564}
(\byear{2023})
\end{barticle}
\endbibitem

%%% 25
\bibitem[\protect\citeauthoryear{Rosenthal et~al.}{2025}]{rosenthal2025rethinking}
\begin{barticle}
\bauthor{\bsnm{Rosenthal}, \binits{J.T.}},
\bauthor{\bsnm{Beecy}, \binits{A.}},
\bauthor{\bsnm{Sabuncu}, \binits{M.R.}}:
\batitle{Rethinking clinical trials for medical ai with dynamic deployments of adaptive systems}.
\bjtitle{npj Digital Medicine}
\bvolume{8}(\bissue{1}),
\bfpage{1}--\blpage{6}
(\byear{2025})
\end{barticle}
\endbibitem

%%% 26
\bibitem[\protect\citeauthoryear{Feng et~al.}{2025}]{feng2025doctoragent}
\begin{botherref}
\oauthor{\bsnm{Feng}, \binits{Y.}},
\oauthor{\bsnm{Wang}, \binits{J.}},
\oauthor{\bsnm{Zhou}, \binits{L.}},
\oauthor{\bsnm{Li}, \binits{Y.}}:
Doctoragent-rl: A multi-agent collaborative reinforcement learning system for multi-turn clinical dialogue.
arXiv preprint arXiv:2505.19630
(2025)
\end{botherref}
\endbibitem

%%% 27
\bibitem[\protect\citeauthoryear{Mehandru et~al.}{2024}]{mehandru2024evaluating}
\begin{barticle}
\bauthor{\bsnm{Mehandru}, \binits{N.}},
\bauthor{\bsnm{Miao}, \binits{B.Y.}},
\bauthor{\bsnm{Almaraz}, \binits{E.R.}},
\bauthor{\bsnm{Sushil}, \binits{M.}},
\bauthor{\bsnm{Butte}, \binits{A.J.}},
\bauthor{\bsnm{Alaa}, \binits{A.}}:
\batitle{Evaluating large language models as agents in the clinic}.
\bjtitle{NPJ digital medicine}
\bvolume{7}(\bissue{1}),
\bfpage{84}
(\byear{2024})
\end{barticle}
\endbibitem

%%% 28
\bibitem[\protect\citeauthoryear{Liu et~al.}{2025}]{liu2025uncertainty}
\begin{botherref}
\oauthor{\bsnm{Liu}, \binits{X.}},
\oauthor{\bsnm{Chen}, \binits{T.}},
\oauthor{\bsnm{Da}, \binits{L.}},
\oauthor{\bsnm{Chen}, \binits{C.}},
\oauthor{\bsnm{Lin}, \binits{Z.}},
\oauthor{\bsnm{Wei}, \binits{H.}}:
Uncertainty quantification and confidence calibration in large language models: A survey.
arXiv preprint arXiv:2503.15850
(2025)
\end{botherref}
\endbibitem

%%% 29
\bibitem[\protect\citeauthoryear{Savage et~al.}{2025}]{savage2025large}
\begin{barticle}
\bauthor{\bsnm{Savage}, \binits{T.}},
\bauthor{\bsnm{Wang}, \binits{J.}},
\bauthor{\bsnm{Gallo}, \binits{R.}},
\bauthor{\bsnm{Boukil}, \binits{A.}},
\bauthor{\bsnm{Patel}, \binits{V.}},
\bauthor{\bsnm{Safavi-Naini}, \binits{S.A.A.}},
\bauthor{\bsnm{Soroush}, \binits{A.}},
\bauthor{\bsnm{Chen}, \binits{J.H.}}:
\batitle{Large language model uncertainty proxies: discrimination and calibration for medical diagnosis and treatment}.
\bjtitle{Journal of the American Medical Informatics Association}
\bvolume{32}(\bissue{1}),
\bfpage{139}--\blpage{149}
(\byear{2025})
\end{barticle}
\endbibitem

%%% 30
\bibitem[\protect\citeauthoryear{Gallifant et~al.}{2025}]{gallifant2025tripod}
\begin{botherref}
\oauthor{\bsnm{Gallifant}, \binits{J.}},
\oauthor{\bsnm{Afshar}, \binits{M.}},
\oauthor{\bsnm{Ameen}, \binits{S.}},
\oauthor{\bsnm{Aphinyanaphongs}, \binits{Y.}},
\oauthor{\bsnm{Chen}, \binits{S.}},
\oauthor{\bsnm{Cacciamani}, \binits{G.}},
\oauthor{\bsnm{Demner-Fushman}, \binits{D.}},
\oauthor{\bsnm{Dligach}, \binits{D.}},
\oauthor{\bsnm{Daneshjou}, \binits{R.}},
\oauthor{\bsnm{Fernandes}, \binits{C.}}, et al.:
The tripod-llm reporting guideline for studies using large language models.
Nature Medicine,
1--10
(2025)
\end{botherref}
\endbibitem

%%% 31
\bibitem[\protect\citeauthoryear{Johnson et~al.}{2023}]{johnson2023mimic}
\begin{botherref}
\oauthor{\bsnm{Johnson}, \binits{A.}},
\oauthor{\bsnm{Bulgarelli}, \binits{L.}},
\oauthor{\bsnm{Pollard}, \binits{T.}},
\oauthor{\bsnm{Horng}, \binits{S.}},
\oauthor{\bsnm{Celi}, \binits{L.A.}},
\oauthor{\bsnm{Mark}, \binits{R.}}:
{MIMIC-IV (version 2.2)}.
\url{https://doi.org/10.13026/6mm1-ek67}.
PhysioNet. RRID:SCR\_007345
(2023)
\end{botherref}
\endbibitem

%%% 32
\bibitem[\protect\citeauthoryear{Singer et~al.}{2016}]{singer2016sepsis3}
\begin{barticle}
\bauthor{\bsnm{Singer}, \binits{M.}},
\bauthor{\bsnm{Deutschman}, \binits{C.S.}},
\bauthor{\bsnm{Seymour}, \binits{C.W.}},
\bauthor{\bsnm{Shankar-Hari}, \binits{M.}},
\bauthor{\bsnm{Annane}, \binits{D.}},
\bauthor{\bsnm{Bauer}, \binits{M.}},
\bauthor{\bsnm{Bellomo}, \binits{R.}},
\bauthor{\bsnm{Bernard}, \binits{G.R.}},
\bauthor{\bsnm{Chiche}, \binits{J.-D.}},
\bauthor{\bsnm{Coopersmith}, \binits{C.M.}},
\bauthor{\bsnm{Hotchkiss}, \binits{R.S.}},
\bauthor{\bsnm{Levy}, \binits{M.M.}},
\bauthor{\bsnm{Marshall}, \binits{J.C.}},
\bauthor{\bsnm{Martin}, \binits{G.S.}},
\bauthor{\bsnm{Opal}, \binits{S.M.}},
\bauthor{\bsnm{Rubenfeld}, \binits{G.D.}},
\bauthor{\bsnm{Poll}, \binits{T.}},
\bauthor{\bsnm{Vincent}, \binits{J.-L.}},
\bauthor{\bsnm{Angus}, \binits{D.C.}}:
\batitle{The third international consensus definitions for sepsis and septic shock (sepsis-3)}.
\bjtitle{JAMA}
\bvolume{315}(\bissue{8}),
\bfpage{801}--\blpage{810}
(\byear{2016})
\doiurl{10.1001/jama.2016.0287}
\end{barticle}
\endbibitem

%%% 33
\bibitem[\protect\citeauthoryear{Yang et~al.}{2025}]{qwen3}
\begin{botherref}
\oauthor{\bsnm{Yang}, \binits{A.}},
\oauthor{\bsnm{Li}, \binits{A.}},
\oauthor{\bsnm{Yang}, \binits{B.}},
\oauthor{\bsnm{Zhang}, \binits{B.}},
\oauthor{\bsnm{Hui}, \binits{B.}},
\oauthor{\bsnm{Zheng}, \binits{B.}},
\oauthor{\bsnm{Yu}, \binits{B.}},
\oauthor{\bsnm{Gao}, \binits{C.}},
\oauthor{\bsnm{Huang}, \binits{C.}},
\oauthor{\bsnm{Lv}, \binits{C.}}, et al.:
Qwen3 technical report.
arXiv preprint arXiv:2505.09388
(2025)
\end{botherref}
\endbibitem

%%% 34
\bibitem[\protect\citeauthoryear{Hu et~al.}{2022}]{hu2022lora}
\begin{barticle}
\bauthor{\bsnm{Hu}, \binits{E.J.}},
\bauthor{\bsnm{Shen}, \binits{Y.}},
\bauthor{\bsnm{Wallis}, \binits{P.}},
\bauthor{\bsnm{Allen-Zhu}, \binits{Z.}},
\bauthor{\bsnm{Li}, \binits{Y.}},
\bauthor{\bsnm{Wang}, \binits{S.}},
\bauthor{\bsnm{Wang}, \binits{L.}},
\bauthor{\bsnm{Chen}, \binits{W.}}, \betal:
\batitle{Lora: Low-rank adaptation of large language models.}
\bjtitle{ICLR}
\bvolume{1}(\bissue{2}),
\bfpage{3}
(\byear{2022})
\end{barticle}
\endbibitem

%%% 35
\bibitem[\protect\citeauthoryear{Schulman et~al.}{2017}]{schulman2017ppo}
\begin{botherref}
\oauthor{\bsnm{Schulman}, \binits{J.}},
\oauthor{\bsnm{Wolski}, \binits{F.}},
\oauthor{\bsnm{Dhariwal}, \binits{P.}},
\oauthor{\bsnm{Radford}, \binits{A.}},
\oauthor{\bsnm{Klimov}, \binits{O.}}:
Proximal policy optimization algorithms.
arXiv preprint arXiv:1707.06347
(2017)
\end{botherref}
\endbibitem

%%% 36
\bibitem[\protect\citeauthoryear{Che et~al.}{2018}]{che2018recurrent}
\begin{barticle}
\bauthor{\bsnm{Che}, \binits{Z.}},
\bauthor{\bsnm{Purushotham}, \binits{S.}},
\bauthor{\bsnm{Cho}, \binits{K.}},
\bauthor{\bsnm{Sontag}, \binits{D.}},
\bauthor{\bsnm{Liu}, \binits{Y.}}:
\batitle{Recurrent neural networks for multivariate time series with missing values}.
\bjtitle{Scientific reports}
\bvolume{8}(\bissue{1}),
\bfpage{6085}
(\byear{2018})
\end{barticle}
\endbibitem

%%% 37
\bibitem[\protect\citeauthoryear{Rajkomar et~al.}{2018}]{rajkomar2018scalable}
\begin{barticle}
\bauthor{\bsnm{Rajkomar}, \binits{A.}},
\bauthor{\bsnm{Oren}, \binits{E.}},
\bauthor{\bsnm{Chen}, \binits{K.}},
\bauthor{\bsnm{Dai}, \binits{A.M.}},
\bauthor{\bsnm{Hajaj}, \binits{N.}},
\bauthor{\bsnm{Hardt}, \binits{M.}},
\bauthor{\bsnm{Liu}, \binits{P.J.}},
\bauthor{\bsnm{Liu}, \binits{X.}},
\bauthor{\bsnm{Marcus}, \binits{J.}},
\bauthor{\bsnm{Sun}, \binits{M.}}, \betal:
\batitle{Scalable and accurate deep learning with electronic health records}.
\bjtitle{NPJ digital medicine}
\bvolume{1}(\bissue{1}),
\bfpage{18}
(\byear{2018})
\end{barticle}
\endbibitem

%%% 38
\bibitem[\protect\citeauthoryear{Futoma et~al.}{2017}]{futoma2017improved}
\begin{bchapter}
\bauthor{\bsnm{Futoma}, \binits{J.}},
\bauthor{\bsnm{Hariharan}, \binits{S.}},
\bauthor{\bsnm{Heller}, \binits{K.}},
\bauthor{\bsnm{Sendak}, \binits{M.}},
\bauthor{\bsnm{Brajer}, \binits{N.}},
\bauthor{\bsnm{Clement}, \binits{M.}},
\bauthor{\bsnm{Bedoya}, \binits{A.}},
\bauthor{\bsnm{O’brien}, \binits{C.}}:
\bctitle{An improved multi-output gaussian process rnn with real-time validation for early sepsis detection}.
In: \bbtitle{Machine Learning for Healthcare Conference},
pp. \bfpage{243}--\blpage{254}
(\byear{2017}).
\bcomment{PMLR}
\end{bchapter}
\endbibitem

%%% 39
\bibitem[\protect\citeauthoryear{Murtaza et~al.}{2023}]{murtaza2023synthetic}
\begin{barticle}
\bauthor{\bsnm{Murtaza}, \binits{H.}},
\bauthor{\bsnm{Ahmed}, \binits{M.}},
\bauthor{\bsnm{Khan}, \binits{N.F.}},
\bauthor{\bsnm{Murtaza}, \binits{G.}},
\bauthor{\bsnm{Zafar}, \binits{S.}},
\bauthor{\bsnm{Bano}, \binits{A.}}:
\batitle{Synthetic data generation: State of the art in health care domain}.
\bjtitle{Computer Science Review}
\bvolume{48},
\bfpage{100546}
(\byear{2023})
\end{barticle}
\endbibitem

%%% 40
\bibitem[\protect\citeauthoryear{Liu et~al.}{2020}]{liu2020reporting}
\begin{barticle}
\bauthor{\bsnm{Liu}, \binits{X.}},
\bauthor{\bsnm{Rivera}, \binits{S.C.}},
\bauthor{\bsnm{Moher}, \binits{D.}},
\bauthor{\bsnm{Calvert}, \binits{M.J.}},
\bauthor{\bsnm{Denniston}, \binits{A.K.}},
\bauthor{\bsnm{Ashrafian}, \binits{H.}},
\bauthor{\bsnm{Beam}, \binits{A.L.}},
\bauthor{\bsnm{Chan}, \binits{A.-W.}},
\bauthor{\bsnm{Collins}, \binits{G.S.}},
\bauthor{\bsnm{Deeks}, \binits{A.D.J.}}, \betal:
\batitle{Reporting guidelines for clinical trial reports for interventions involving artificial intelligence: the consort-ai extension}.
\bjtitle{The Lancet Digital Health}
\bvolume{2}(\bissue{10}),
\bfpage{537}--\blpage{548}
(\byear{2020})
\end{barticle}
\endbibitem

%%% 41
\bibitem[\protect\citeauthoryear{Esteban et~al.}{2017}]{esteban2017real}
\begin{botherref}
\oauthor{\bsnm{Esteban}, \binits{C.}},
\oauthor{\bsnm{Hyland}, \binits{S.L.}},
\oauthor{\bsnm{R{\"a}tsch}, \binits{G.}}:
Real-valued (medical) time series generation with recurrent conditional gans.
arXiv preprint arXiv:1706.02633
(2017)
\end{botherref}
\endbibitem

%%% 42
\bibitem[\protect\citeauthoryear{Mangrulkar et~al.}{2022}]{peft}
\begin{botherref}
\oauthor{\bsnm{Mangrulkar}, \binits{S.}},
\oauthor{\bsnm{Gugger}, \binits{S.}},
\oauthor{\bsnm{Debut}, \binits{L.}},
\oauthor{\bsnm{Belkada}, \binits{Y.}},
\oauthor{\bsnm{Paul}, \binits{S.}},
\oauthor{\bsnm{Bossan}, \binits{B.}}:
{PEFT}: State-of-the-art Parameter-Efficient Fine-Tuning methods.
\url{https://github.com/huggingface/peft}
(2022)
\end{botherref}
\endbibitem

%%% 43
\bibitem[\protect\citeauthoryear{{OpenAI}}{2025}]{openai_gpt4_1}
\begin{botherref}
\oauthor{\bsnm{{OpenAI}}}:
Introducing GPT-4.1 in the API.
\url{https://openai.com/index/gpt-4-1/}
(2025)
\end{botherref}
\endbibitem

%%% 44
\bibitem[\protect\citeauthoryear{{Anthropic}}{2025}]{anthropic_claude_sonnet_4}
\begin{botherref}
\oauthor{\bsnm{{Anthropic}}}:
Introducing Claude 4.
\url{https://www.anthropic.com/news/claude-4}
(2025)
\end{botherref}
\endbibitem

%%% 45
\bibitem[\protect\citeauthoryear{Yu et~al.}{2025}]{yu2025finemedlm}
\begin{botherref}
\oauthor{\bsnm{Yu}, \binits{H.}},
\oauthor{\bsnm{Cheng}, \binits{T.}},
\oauthor{\bsnm{Cheng}, \binits{Y.}},
\oauthor{\bsnm{Feng}, \binits{R.}}:
Finemedlm-o1: Enhancing the medical reasoning ability of llm from supervised fine-tuning to test-time training.
arXiv preprint arXiv:2501.09213
(2025)
\end{botherref}
\endbibitem

%%% 46
\bibitem[\protect\citeauthoryear{Chen et~al.}{2024}]{chen2024huatuogpt}
\begin{botherref}
\oauthor{\bsnm{Chen}, \binits{J.}},
\oauthor{\bsnm{Cai}, \binits{Z.}},
\oauthor{\bsnm{Ji}, \binits{K.}},
\oauthor{\bsnm{Wang}, \binits{X.}},
\oauthor{\bsnm{Liu}, \binits{W.}},
\oauthor{\bsnm{Wang}, \binits{R.}},
\oauthor{\bsnm{Hou}, \binits{J.}},
\oauthor{\bsnm{Wang}, \binits{B.}}:
Huatuogpt-o1, towards medical complex reasoning with llms.
arXiv preprint arXiv:2412.18925
(2024)
\end{botherref}
\endbibitem

%%% 47
\bibitem[\protect\citeauthoryear{Cai et~al.}{2024}]{cai2024internlm2}
\begin{botherref}
\oauthor{\bsnm{Cai}, \binits{Z.}},
\oauthor{\bsnm{Cao}, \binits{M.}},
\oauthor{\bsnm{Chen}, \binits{H.}},
\oauthor{\bsnm{Chen}, \binits{K.}},
\oauthor{\bsnm{Chen}, \binits{K.}},
\oauthor{\bsnm{Chen}, \binits{X.}},
\oauthor{\bsnm{Chen}, \binits{X.}},
\oauthor{\bsnm{Chen}, \binits{Z.}},
\oauthor{\bsnm{Chen}, \binits{Z.}},
\oauthor{\bsnm{Chu}, \binits{P.}},
\oauthor{\bsnm{Dong}, \binits{X.}},
\oauthor{\bsnm{Duan}, \binits{H.}},
\oauthor{\bsnm{Fan}, \binits{Q.}},
\oauthor{\bsnm{Fei}, \binits{Z.}},
\oauthor{\bsnm{Gao}, \binits{Y.}},
\oauthor{\bsnm{Ge}, \binits{J.}},
\oauthor{\bsnm{Gu}, \binits{C.}},
\oauthor{\bsnm{Gu}, \binits{Y.}},
\oauthor{\bsnm{Gui}, \binits{T.}},
\oauthor{\bsnm{Guo}, \binits{A.}},
\oauthor{\bsnm{Guo}, \binits{Q.}},
\oauthor{\bsnm{He}, \binits{C.}},
\oauthor{\bsnm{Hu}, \binits{Y.}},
\oauthor{\bsnm{Huang}, \binits{T.}},
\oauthor{\bsnm{Jiang}, \binits{T.}},
\oauthor{\bsnm{Jiao}, \binits{P.}},
\oauthor{\bsnm{Jin}, \binits{Z.}},
\oauthor{\bsnm{Lei}, \binits{Z.}},
\oauthor{\bsnm{Li}, \binits{J.}},
\oauthor{\bsnm{Li}, \binits{J.}},
\oauthor{\bsnm{Li}, \binits{L.}},
\oauthor{\bsnm{Li}, \binits{S.}},
\oauthor{\bsnm{Li}, \binits{W.}},
\oauthor{\bsnm{Li}, \binits{Y.}},
\oauthor{\bsnm{Liu}, \binits{H.}},
\oauthor{\bsnm{Liu}, \binits{J.}},
\oauthor{\bsnm{Hong}, \binits{J.}},
\oauthor{\bsnm{Liu}, \binits{K.}},
\oauthor{\bsnm{Liu}, \binits{K.}},
\oauthor{\bsnm{Liu}, \binits{X.}},
\oauthor{\bsnm{Lv}, \binits{C.}},
\oauthor{\bsnm{Lv}, \binits{H.}},
\oauthor{\bsnm{Lv}, \binits{K.}},
\oauthor{\bsnm{Ma}, \binits{L.}},
\oauthor{\bsnm{Ma}, \binits{R.}},
\oauthor{\bsnm{Ma}, \binits{Z.}},
\oauthor{\bsnm{Ning}, \binits{W.}},
\oauthor{\bsnm{Ouyang}, \binits{L.}},
\oauthor{\bsnm{Qiu}, \binits{J.}},
\oauthor{\bsnm{Qu}, \binits{Y.}},
\oauthor{\bsnm{Shang}, \binits{F.}},
\oauthor{\bsnm{Shao}, \binits{Y.}},
\oauthor{\bsnm{Song}, \binits{D.}},
\oauthor{\bsnm{Song}, \binits{Z.}},
\oauthor{\bsnm{Sui}, \binits{Z.}},
\oauthor{\bsnm{Sun}, \binits{P.}},
\oauthor{\bsnm{Sun}, \binits{Y.}},
\oauthor{\bsnm{Tang}, \binits{H.}},
\oauthor{\bsnm{Wang}, \binits{B.}},
\oauthor{\bsnm{Wang}, \binits{G.}},
\oauthor{\bsnm{Wang}, \binits{J.}},
\oauthor{\bsnm{Wang}, \binits{J.}},
\oauthor{\bsnm{Wang}, \binits{R.}},
\oauthor{\bsnm{Wang}, \binits{Y.}},
\oauthor{\bsnm{Wang}, \binits{Z.}},
\oauthor{\bsnm{Wei}, \binits{X.}},
\oauthor{\bsnm{Weng}, \binits{Q.}},
\oauthor{\bsnm{Wu}, \binits{F.}},
\oauthor{\bsnm{Xiong}, \binits{Y.}},
\oauthor{\bsnm{Xu}, \binits{C.}},
\oauthor{\bsnm{Xu}, \binits{R.}},
\oauthor{\bsnm{Yan}, \binits{H.}},
\oauthor{\bsnm{Yan}, \binits{Y.}},
\oauthor{\bsnm{Yang}, \binits{X.}},
\oauthor{\bsnm{Ye}, \binits{H.}},
\oauthor{\bsnm{Ying}, \binits{H.}},
\oauthor{\bsnm{Yu}, \binits{J.}},
\oauthor{\bsnm{Yu}, \binits{J.}},
\oauthor{\bsnm{Zang}, \binits{Y.}},
\oauthor{\bsnm{Zhang}, \binits{C.}},
\oauthor{\bsnm{Zhang}, \binits{L.}},
\oauthor{\bsnm{Zhang}, \binits{P.}},
\oauthor{\bsnm{Zhang}, \binits{P.}},
\oauthor{\bsnm{Zhang}, \binits{R.}},
\oauthor{\bsnm{Zhang}, \binits{S.}},
\oauthor{\bsnm{Zhang}, \binits{S.}},
\oauthor{\bsnm{Zhang}, \binits{W.}},
\oauthor{\bsnm{Zhang}, \binits{W.}},
\oauthor{\bsnm{Zhang}, \binits{X.}},
\oauthor{\bsnm{Zhang}, \binits{X.}},
\oauthor{\bsnm{Zhao}, \binits{H.}},
\oauthor{\bsnm{Zhao}, \binits{Q.}},
\oauthor{\bsnm{Zhao}, \binits{X.}},
\oauthor{\bsnm{Zhou}, \binits{F.}},
\oauthor{\bsnm{Zhou}, \binits{Z.}},
\oauthor{\bsnm{Zhuo}, \binits{J.}},
\oauthor{\bsnm{Zou}, \binits{Y.}},
\oauthor{\bsnm{Qiu}, \binits{X.}},
\oauthor{\bsnm{Qiao}, \binits{Y.}},
\oauthor{\bsnm{Lin}, \binits{D.}}:
InternLM2 Technical Report
(2024)
\end{botherref}
\endbibitem

%%% 48
\bibitem[\protect\citeauthoryear{Grattafiori et~al.}{2024}]{grattafiori2024llama}
\begin{botherref}
\oauthor{\bsnm{Grattafiori}, \binits{A.}},
\oauthor{\bsnm{Dubey}, \binits{A.}},
\oauthor{\bsnm{Jauhri}, \binits{A.}},
\oauthor{\bsnm{Pandey}, \binits{A.}},
\oauthor{\bsnm{Kadian}, \binits{A.}},
\oauthor{\bsnm{Al-Dahle}, \binits{A.}},
\oauthor{\bsnm{Letman}, \binits{A.}},
\oauthor{\bsnm{Mathur}, \binits{A.}},
\oauthor{\bsnm{Schelten}, \binits{A.}},
\oauthor{\bsnm{Vaughan}, \binits{A.}}, et al.:
The llama 3 herd of models.
arXiv preprint arXiv:2407.21783
(2024)
\end{botherref}
\endbibitem

\end{thebibliography}

\end{document}